\pdfoutput=1
\documentclass[pdftex]{article}

\newcommand{\ie}{\textit{i.e.}}
\newcommand{\etal}{\textit{et al.}}
\newcommand{\eg}{\textit{e.g.}}

\usepackage[nonatbib,final]{neurips_2024}
\definecolor{linkc}{rgb}{0, 0.44, 0.74}
\definecolor{eqc}{rgb}{0.8, 0, 0}

\usepackage{xspace}
\usepackage[utf8]{inputenc} 
\usepackage[T1]{fontenc}    
\usepackage{hyperref}
\usepackage{url}            
\usepackage{booktabs}       
\usepackage{amsfonts}       
\usepackage{nicefrac}       
\usepackage{microtype}      
\usepackage{xcolor}         
\usepackage{subfigure}
\usepackage{graphicx}
\usepackage{caption}

\usepackage{textcomp}
\usepackage{pifont}
\usepackage{endnotes}
\usepackage{multirow}
\usepackage[ruled,vlined]{algorithm2e}
\usepackage{colortbl}
\usepackage{amsmath}
\usepackage{float}

\usepackage[utf8]{inputenc} 
\usepackage[T1]{fontenc}    

\title{Masked Pre-training Enables Universal\\Zero-shot Denoiser}

\author{%
Xiaoxiao Ma{$^{1}$\thanks{Equal contribution.}} ~ Zhixiang Wei{$^{1*}$} ~ Yi Jin{$^{1*}$} ~ Pengyang Ling{$^{1,2}$} ~ Tianle Liu{$^{1}$} ~ \\ \textbf{Ben Wang}{$^{1}$} ~ \textbf{Junkang Dai}{$^{1}$} ~ \textbf{Huaian Chen}{$^{1}$\footnotemark[2]\thanks{Corresponding author.}}\\
\normalsize
$^{1}$\	University of Science and Technology of China ~~ $^{2}$\,Shanghai AI Laboratory \\
\normalsize
{\tt\small \{xiao\_xiao,zhixiangwei,lpyang27,tleliu,wblzgrsn,junkangdai,anchen\}@mail.ustc.edu.cn} \\
{\tt\small \{jinyi08\}@ustc.edu.cn}
}

\begin{document}

\maketitle

\vspace{-3mm}
\begin{figure}[h]
\setlength{\abovecaptionskip}{0.2cm}
\setlength{\belowcaptionskip}{0.2cm}
  \centering
   \includegraphics[width=0.95\linewidth]{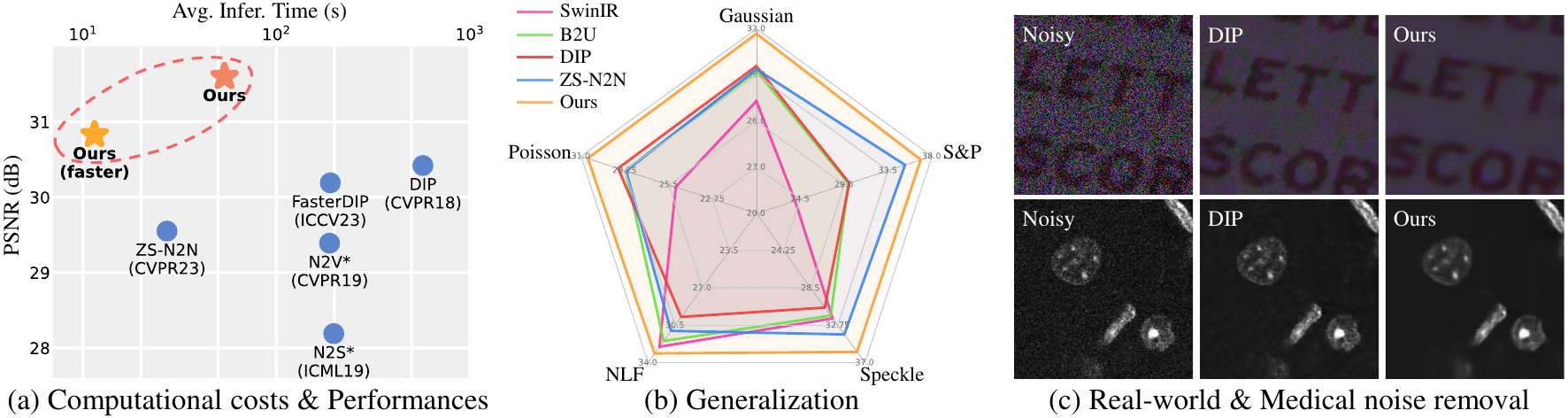}

   \caption{(a) Ours surpasses current zero-shot methods with reduced inference time (on CSet with Gaussian $\sigma$=25, see Sec.~\ref{sec:awgn&poisson}). (b) It shows better generalization across different noise types than current zero-shot \& supervised/unsupervised methods (Sec.~\ref{sec:syn_general}). (c) And can remove spatial correlated real-world noise, results are from SIDD benchmark~\cite{abdelhamed2018sidd} and FMD~\cite{zhang2019fmd} (Sec.~\ref{sec:real_noise}, Sec.~\ref{sec:medical_noise}).}
   \label{fig:teaser}
\end{figure}
\begin{abstract}
In this work, we observe that model trained on vast general images via masking strategy, has been naturally embedded with their distribution knowledge, thus spontaneously attains the underlying potential for strong image denoising.
Based on this observation, we propose a novel zero-shot denoising paradigm, i.e., \textbf{M}asked \textbf{P}re-train then \textbf{I}terative fill (\textbf{MPI}).
MPI first trains model via masking and then employs pre-trained weight for high-quality zero-shot image denoising on a single noisy image.
Concretely, MPI comprises two key procedures:
\textbf{1) Masked Pre-training} involves training model to reconstruct massive natural images with random masking for generalizable representations, gathering the potential for valid zero-shot denoising on images with varying noise degradation and even in distinct image types.
\textbf{2) Iterative filling} exploits pre-trained knowledge for effective zero-shot denoising. It iteratively optimizes the image by leveraging pre-trained weights, focusing on alternate reconstruction of different image parts, and gradually assembles fully denoised image within limited number of iterations.
Comprehensive experiments across various noisy scenarios underscore the notable advances of MPI over previous approaches with a marked reduction in inference time. Code available at~\url{https://github.com/krennic999/MPI}.
\end{abstract}
\section{Introduction}
\label{sec:introduction}
Image denoising~\cite{dabov2007bm3d,maggioni2012bm4d}, as a branch of image restoration, has been the subject of extensive exploration. The prevalent approach to restore noise-degraded images is learning from multiple noisy instances. Nonetheless, both supervised learning from noisy-clean pairs~\cite{zhang2017dncnn,zhang2018ffdnet,liang2021swinir} and unsupervised training~\cite{lehtinen2018noise2noise,du2020invariant,huang2021neighbor2neighbor} necessitate the collection of additional noisy datasets. Moreover, such methods may foster dependencies on specific patterns or intensities of training noise, hindering their performance in unfamiliar noise situations~\cite{chen2023med,chen2023maskdenoising}.

As an alternative, zero-shot approaches~\cite{Ulyanov2018dip,quan2020self2self,cheng2023scoreprior} attempt to train network on a single noisy image for denoised output, negating the need for additional noisy data collection. Dedicated to obviating concerns about generalization issues, these techniques include blind-spot networks that reconstruct from corrupted inputs~\cite{krull2019noise2void,batson2019noise2self}, DIPs~\cite{Ulyanov2018dip,heckel2018deep-decoder,liu2023fasterdip,jo2021rethinkingdip,shi2022measuringdip} which exploit the characteristics of deep networks to learn the mapping from random noise to noisy images, as well as sub-sample based strategies~\cite{Mansour2023zs-n2n,lequyer2021noise2fast} utilize spatial correlations to generate training pairs from sub-sampled instances.

However, current zero-shot methods train new networks from scratch for each noisy image, which presents two major issues:
1) Despite success in current zero-shot approaches rely on regularization or designed priors such as noise perturbations~\cite{Ulyanov2018dip}, under-parameterized networks~\cite{heckel2018deep-decoder,Mansour2023zs-n2n}, dropout-ensemble~\cite{quan2020self2self} and blind-spot networks~\cite{krull2019noise2void}, the limited information from a single image to train network often lead to overly blurred content, noise artifacts or sub-optimal quality. Several methods tend to rely on known noise distribution~\cite{dabov2007bm3d,Makitalo2010possiondenoise,jo2021rethinkingdip} for more information, but their applicability is limited.
2) Training a new network from scratch for each noisy image is time-consuming. Existing zero-shot methods typically require several minutes~\cite{Ulyanov2018dip} or more~\cite{quan2020self2self}. And attempts at faster zero-shot denoising~\cite{Mansour2023zs-n2n,lequyer2021noise2fast} often compromise on performance.

Compared to previous zero-shot approaches, learning the feature distribution from vast natural images offers a more intuitive approach. This is grounded in two considerations: Real natural images are both abundant and readily available, and despite variations in noise patterns, many natural images share common characteristics~\cite{mittal2012niqe}.
We seek to enhance zero-shot denoising with minimal reliance on pre-defined priors or regularization, aiming for a better startpoint for various noise patterns instead of from scratch.
To this end, we delve into the potential of masked image modelling~\cite{he2022mae,xie2022simmim} on natural images with no assumptions about noisy patterns and intensities~\cite{deng2009imagenet}. 
Specifically, we make the following observation: combined with a simple ensemble operation, \textbf{a masked pre-trained model can naturally denoise images with unseen noise degradation}.

Building upon above observation, we introduce a zero-shot denoising paradigm, \ie, Masked Pre-train then Iterative fill (MPI). MPI first pre-trains a model on ImageNet with pixel-wise masking strategy, then the pre-trained model is optimized on a single image with unseen noise for denoised prediction in zero-shot inference stage.
The optimization goal in inference is designed to predict masked regions, and only predictions of masked areas are preserved for denoised prediction, thereby minimizing the gap between pre-training and inference.
The pre-trained weights provide more generic knowledge, preventing premature over-fitting during inference and reducing the need for strong regularization.
We are able to handle a wider range of noise scenarios with less information about noise patterns or intensities.
Remarkably, we find that extracted representation can even generalize to medical images that distinctly different from natural ones~\cite{zhang2019fmd}.
It also offers a better startpoint than scratch training, enabling high-quality denoising around 10 seconds, underscoring the potential of our method in practical application. 
The main contributions of this paper are as follows:

\setlength{\leftmargini}{2em}
\begin{itemize}
    \item We introduce a novel zero-shot denoising paradigm, \ie, Masked Pre-train then Iterative fill (MPI), which introduces masked pre-training in this context for the first time,
    simultaneously improving both image quality and inference speed on unseen noisy images.
    \item We develop a pre-training scheme with pixel-wise random masks to capture distribution knowledge of natural images. Based on pre-trained knowledge, we propose iterative filling for zero-shot inference on a specific noisy image. This process is optimized using pre-trained weights, and focuses on alternatively reconstruct different parts of noisy image, predictions in iterations are sequentially assembled for high-quality denoised output with efficiency.
    \item Extensive experiments demonstrate MPI's superiority, efficiency and robustness in diverse noisy scenarios. In nutshell, MPI achieves significant performance gains across various noise types with reduced inference time, highlighting its potential for practical applications.
\end{itemize}

\section{Methods}
\label{sec:methods}

In Sec.~\ref{sec:motivation}, we first investigate the properties of models trained with masking, proving that models trained with masking strategy can learn representations beneficial for denoising. Our observation lead us to propose a zero-shot denoising paradigm that includes pre-training (Sec.~\ref{sec:mim_pretrain}) and iterative optimizing (Sec.~\ref{sec:zeroshot_denoise}). We further illustrate how to remove spatially correlated real noise in Sec.~\ref{sec:method_realnoise}.


\noindent
\begin{minipage}[t]{0.58\columnwidth}
\vspace{-2mm}
\setlength{\abovecaptionskip}{0.1cm}
\setlength{\belowcaptionskip}{0.1cm}
  \centering
   \includegraphics[width=1\linewidth]{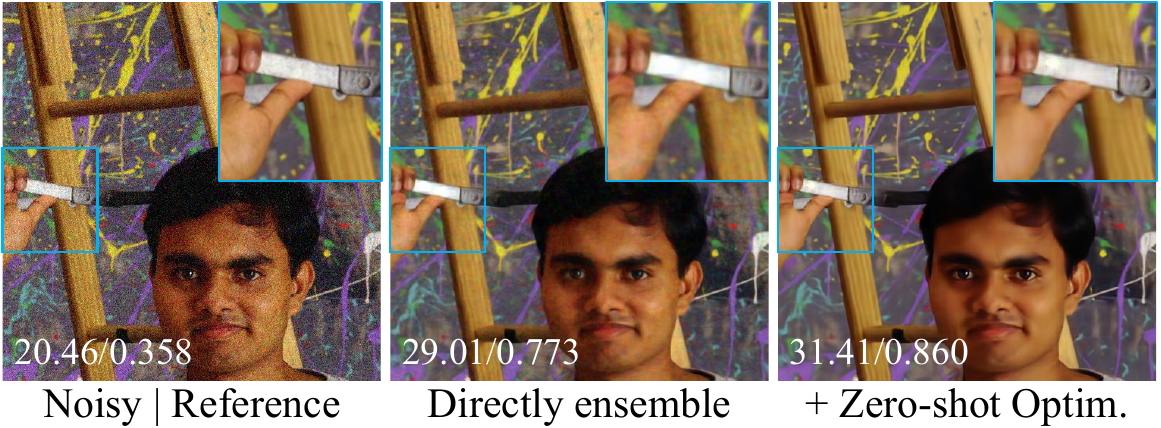}
   \captionof{figure}{Example of model trained on ImageNet with 70\% pixel-wise masking, denoised image is obtained by directly ensemble of predictions from fixed pre-trained weights (``Directly ensemble''), its performance can be further improved with iterative filling (``+Zero-shot Optim.'').}
   \vspace{-0.5cm}
   \label{fig_motivation}
\end{minipage}%
\hfill
\begin{minipage}[t]{0.38\columnwidth}
\vspace{-3mm}
\setlength{\abovecaptionskip}{0.1cm}
\setlength{\belowcaptionskip}{0.1cm}
\centering
  \centering
   \includegraphics[width=0.95\linewidth]{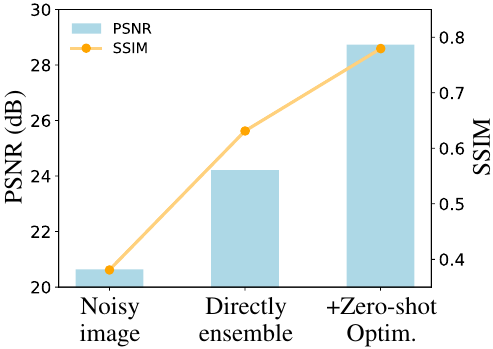}
   \captionof{figure}{Evaluation on an ImageNet subset shows pre-trained model's inherent denoising ability, but performance limited without optimization.}
   \vspace{-0.5cm}
   \label{fig_psnr_pretrain}
\end{minipage}
\vspace{4mm}

\subsection{Motivation}
\label{sec:motivation}
Masked Image Modeling~\cite{he2022mae,xie2022simmim,bao2021beit} has significantly advanced computer vision by training on vast natural image sets to grasp their knowledge distributions. It shows great potential applicability under diverse scenarios and have been proven beneficial for high-level downstream tasks~\cite{mao2023masked_motion_predict,zhai2023masked_class_incremental_learning}.

To further explore its capability in denoising, we train a model on natural images with pixel-wise random masks (for details, see Sec.~\ref{sec:mim_pretrain}) and assess its performance against a target image with unseen noise distribution.
Surprisingly, we observe that a simple average of predictions from a fixed-state trained model can denoise on unseen noise, as shown in Fig.~\ref{fig_psnr_pretrain}, sometimes achieve remarkably good performance, as an example is presented in Fig.~\ref{fig_motivation}. This observation suggests that \textbf{a masked pre-trained model can serve as a natural image denoiser.}
However, artifacts exist in the results, which can be attributed to lack of knowledge about specific degradation patterns in the target image.

\begin{figure*}[t]
\setlength{\abovecaptionskip}{0.1cm}
\setlength{\belowcaptionskip}{0.1cm}
\begin{center}
\includegraphics[width=0.95\textwidth]{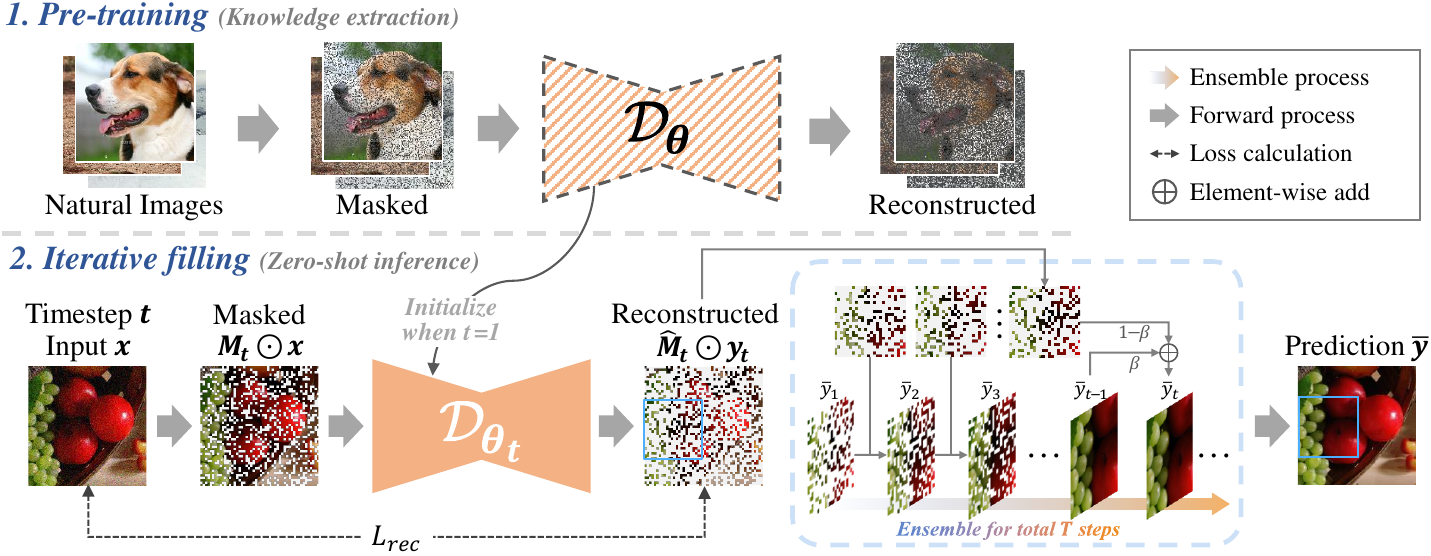}
\end{center}
\caption{
An overview of the proposed MPI paradigm consisting Masked Pre-training and Iterative filling. During pre-training $\mathcal{D}_\theta(\cdot)$ learns to reconstruct masked natural images. And the pre-trained weights $\theta$ are saved for zero-shot denoise, \ie, Iterative filling, to denoise a specific noisy image $x$. During zero-shot inference, network is initialized with pre-trained weights $\theta$, then the weights are further optimized on $x$ for $T$ steps, results from $t$-th ($t$=$1,2,\ldots,T$-$1$) optimizing steps are gathered to obtain final denoised prediction $\overline{y}$. Compared to current zero-shot methods, just adding one more step to load a pre-trained model enables faster and high-quality zero-shot denoising.
}
\vspace{-0.3cm}
\label{fig_framework}
\end{figure*}

Drawing on prior insight, we develop an efficient zero-shot denoising pipeline, leveraging pre-trained knowledge by incorporating noise characteristics from a single noisy image (Fig.~\ref{fig_framework}), \ie, Masked Pre-train then Iterative fill.
The model is firstly pre-trained with random masks $M$ and corresponding element-wise negation $\hat{M}$ to acquire natural image distributions, formulated as:
\begin{equation}
\label{equ_motivation_pretrain}
\mathop{\arg\max}\limits_{\theta} p(I\odot \hat{M}|I\odot M; \theta),
\end{equation}
for $I$ indicates natural image without any degradation priors, typically sourced from extensive datasets (\eg ImageNet~\cite{deng2009imagenet}). We use element-wise multiplication ($\odot$).
For denoising on specific noisy image $x$, pre-trained parameter $\theta$ is loaded and further optimized with known $x$ from $t$=1 to $t$=$T$ for $T$ iterations, and predictions are aggregated for final prediction $\overline{y}$:
\begin{equation}
\label{equ_motivation_ensemble}
\overline{y}=Ensemble\{\mathcal{D}_{\theta_t}(x)\}_{t=1}^{T},
\end{equation}
where $\mathcal{D}_{\theta_t}(\cdot)$ is network parameterized by $\theta_t$, optimized from pre-trained $\theta$.
Masked Pre-training process is detailed in Sec.~\ref{sec:mim_pretrain} and Ensemble in Sec.~\ref{sec:zeroshot_denoise}.

\subsection{Masked Pre-training}
\label{sec:mim_pretrain}


\noindent\textbf{Masking strategy.}
Given the distinct requirements between low-level and high-level tasks in ``semantics''~\cite{liu2021semantics_lowlevel}, we implement specialized masking strategy to achieve finer-grained image representations, \ie, a pixel-wise masking strategy.
Specifically, given an input image $I\in \mathbb R^{H\times W\times C}$ divided into random patches of size 1, a subset of them are randomly replaced by mask token with probability $p$ (for further discussion of $p$, see Sec.~\ref{sec:ablation}).
When the mask token is set to 0, the masked image $M\odot I$ with random mask $M\in \mathbb R^{H\times W\times C}$ corresponds to a bernoulli sampling of the input image $I$. For each element $M_{[k]}$ in $M$, we have:
\begin{equation}
M_{[k]} = \left\{ \begin{array}{ll}
0, & with~prob.~~ p; \\
1, & with~prob.~~ 1-p.
\end{array} \right.
\end{equation}

\noindent\textbf{Pre-training scheme.}
During pre-training, the network $\mathcal{D}_\theta(\cdot)$  is trained to learn recovering natural image $I$ itself with random mask $M$:
\begin{equation}
\tilde{I}=\mathcal{D}_\theta(M\odot I).
\end{equation}
We set the same optimization strategy outlined in~\cite{xie2022simmim}, focusing loss computation on masked prediction areas $\tilde{I}$. This directs network efforts towards reconstructing these specific regions, with the reconstruction loss denoted as $L_{rec}$:
\begin{equation}
L_{rec}(\tilde{I},I)=\left\|\hat{M}\odot\tilde{I}-\hat{M}\odot I\right\|_2.
\label{equ_pretrain_optim}
\end{equation}
The Mean Squared Error (MSE) loss is adopted to learn relatively smoother representation.
For the architecture of network $\mathcal{D}(\cdot)$, we employ the same U-shaped hourglass architecture as in DIP~\cite{Ulyanov2018dip}, which has been proven a powerful zero-shot denoising architecture~\cite{liu2023fasterdip}.
Furthermore, its relatively small parameter configuration enables accelerated training, alleviating potential inference computational costs and rendering it more appropriate for zero-shot denoising tasks.


\vspace{2mm}
\begin{algorithm}[H]
    \KwIn{Noisy image $x$, 
    pre-trained parameter $\theta$, network $\mathcal{D}(\cdot)$,
    exponential weight $\beta$, masking ratio $p$.}
    \KwOut{denoised ensemble $\overline{y}$ from predictions of iteration $\{y_t\}$.}
    load pre-trained parameter $\theta$ for $\mathcal{D}(\cdot)$ as $\theta_1$\\
    initialize $\overline{y}$\\
    \For{$t$ from $1$ to $T$}
    {
    generate random mask $M_t$ with mask ratio $p$\\
    $y_t=\mathcal{D}_{\theta_{t}}(M_t\odot x)$\\
    $\hat{M_t}=\neg M_t$\\
    $\theta_{t+1}=\theta_{t}-\nabla_\theta \left\|\hat{M_t}\odot y_t-\hat{M_t}\odot x\right\|_2$\\
    $\overline{y} \leftarrow \hat{M_t} \odot
    (\beta\cdot \overline{y}+(1-\beta)\cdot y_t)+M_t\odot \overline{y}$
    }
    \caption{\textbf{Iterative filling.} Pipeline designed to leverage pre-trained representation $\theta$ for zero-shot denoising.}
    \label{alg1}
    \Return{$\overline{y}$}
\end{algorithm}
\vspace{-0.5mm} 

\subsection{Iterative Filling}
\label{sec:zeroshot_denoise}
\noindent\textbf{Overall design.}
As observed in Sec.~\ref{sec:motivation}, an iterative optimization process is designed to leverage pre-trained knowledge for zero-shot denoising.
Unlike other MIM approaches~\cite{he2022mae,xie2022simmim} that fine-tune with entire images as input, since only one noisy image is accessible, we employ a self-supervised manner to learn the mapping from a noisy image to itself.
However, this direct self-mapping approach introduces significant gap between the zero-shot inference stage and pre-training stage and lacks constraints for learning a noise identity mapping.

Considering above challenges, we retain the same masking strategy in Sec.~\ref{sec:mim_pretrain} for both input and loss computation, \ie, network still learns to reconstruct masked regions, but from single noisy image rather than natural images. This leads to a pixel-based iterative refinement process, which resembles mechanism of blind-spot networks~\cite{krull2019noise2void}. Specifically, for input noisy image $x$, random mask $M_t$ and its element-wise negation $\hat{M_t}$ in $t$-th iteration, prediction $y_t$ and aggregated result $\overline{y}$ can be derived:
\begin{subequations}
\begin{align}
y_t &= \mathcal{D}_{\theta_t}(M_t\odot x); \label{equ_ensemble_recon}\\
\overline{y} &= \sum_{t}{a_t\cdot y_t\odot \hat{M_t}}, \label{equ_ensemble_overview}
\end{align}
\end{subequations}
\noindent where $\theta_t$ denotes network parameter at iteration $t$, $a_t$ is corresponding coefficient where $\sum_{t}{a_t}=1$. The optimization objective at each iteration is as follows:
\begin{figure*}[t]
\setlength{\abovecaptionskip}{0.1cm}
\setlength{\belowcaptionskip}{0.1cm}
  \centering
   \includegraphics[width=1\linewidth]{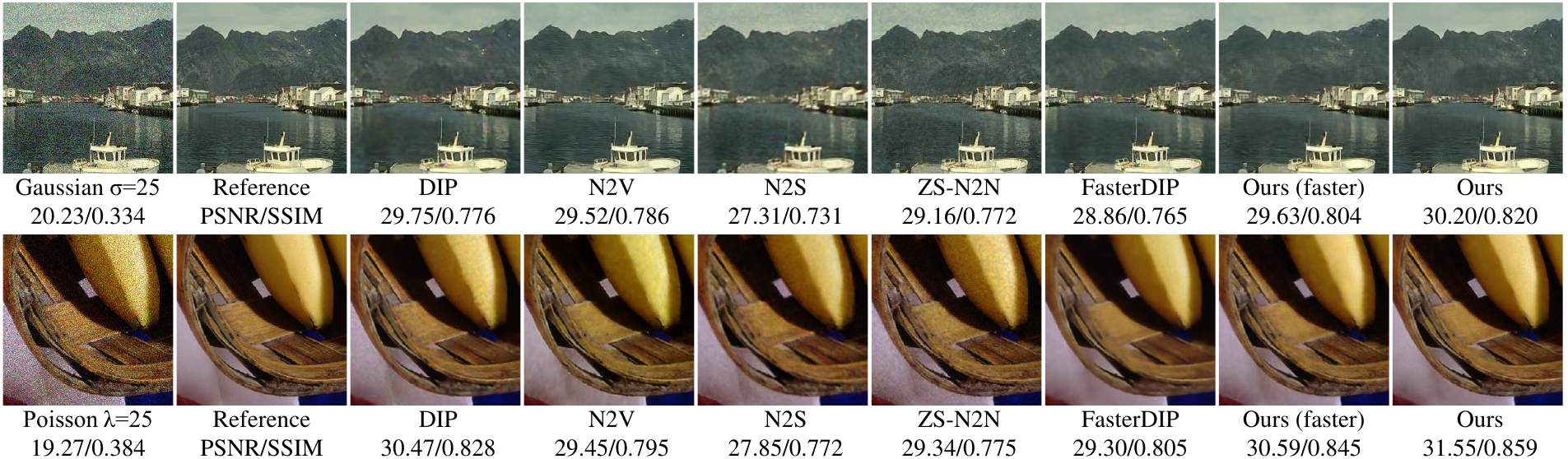}
   \caption{Qualitative denoising results on Gaussian and Poisson noise. The quantitative PSNR/SSIM results are provided underneath. Noisy patches are from CBSD-44 and McMaster-14, respectively. Best viewed in color (zoom-in for better comparison).}
   \vspace{-0.5cm}
   \label{fig_gauss&poiss}
\end{figure*}
\begin{table*}[t]
\centering
\setlength{\abovecaptionskip}{0.1cm}
\setlength{\belowcaptionskip}{0.1cm}
\caption{Quantitative comparison on CSet, McMaster \& CBSD for \textbf{Gaussian noise removal}. Best results \textbf{highlighted} and second \underline{underlined}. See Supp. for poisson noise removal.}
\label{table_gauss}
\setlength{\tabcolsep}{2mm}{
\resizebox{1\columnwidth}{!}{%
\begin{tabular}{ccccccccc}
\toprule
 &
  $\sigma$ &
  DIP~\cite{Ulyanov2018dip} &
  N2V*~\cite{krull2019noise2void} &
  N2S*~\cite{batson2019noise2self} &
  ZS-N2N~\cite{Mansour2023zs-n2n} &
  FasterDIP~\cite{liu2023fasterdip} &
  Ours (faster) &
  Ours \\ \midrule
 &
  10 &
  32.05/0.829 &
  31.55/0.885 &
  28.04/0.819 &
  \underline{33.87}/0.883 &
  31.59/0.815 &
  \cellcolor[HTML]{EFEFEF}33.82/\underline{0.889} &
  \cellcolor[HTML]{EFEFEF}\textbf{34.91/0.909} \\
 &
  25 &
  30.42/0.795 &
  29.39/0.814 &
  28.19/0.777 &
  29.55/0.765 &
  30.19/0.766 &
  \cellcolor[HTML]{EFEFEF}\underline{30.83/0.824} &
  \cellcolor[HTML]{EFEFEF}\textbf{31.61/0.841} \\
\multirow{-3}{*}{CSet~\cite{dabov2007bm3d}} &
  50 &
  24.73/0.533 &
  27.35/0.694 &
  26.62/0.699 &
  26.10/0.624 &
  26.09/0.669 &
  \cellcolor[HTML]{EFEFEF}\underline{28.14}/\textbf{0.715} &
  \cellcolor[HTML]{EFEFEF}\textbf{28.26}/\underline{0.710} \\ \midrule
 &
  10 &
  32.48/0.878 &
  30.98/0.877 &
  28.61/0.839 &
  34.19/0.908 &
  31.48/0.842 &
  \cellcolor[HTML]{EFEFEF}\underline{34.35/0.921} &
  \cellcolor[HTML]{EFEFEF}\textbf{35.46/0.937} \\
 &
  25 &
  \underline{31.07}/0.856 &
  29.11/0.833 &
  27.59/0.776 &
  29.37/0.786 &
  29.47/0.794 &
  \cellcolor[HTML]{EFEFEF}30.99/\underline{0.862} &
  \cellcolor[HTML]{EFEFEF}\textbf{31.90/0.879} \\
\multirow{-3}{*}{McMaster~\cite{ zhang2011mcmaster}} &
  50 &
  25.72/0.639 &
  24.65/0.676 &
  24.89/0.673 &
  25.82/0.634 &
  24.75/0.663 &
  \cellcolor[HTML]{EFEFEF}\underline{28.15}/\textbf{0.779} &
  \cellcolor[HTML]{EFEFEF}\textbf{28.37}/\underline{0.770} \\ \midrule
 &
  10 &
  31.18/0.865 &
  31.18/0.918 &
  28.17/0.853 &
  33.73/0.923 &
  30.89/0.857 &
  \cellcolor[HTML]{EFEFEF}\underline{34.20/0.935} &
  \cellcolor[HTML]{EFEFEF}\textbf{35.14/0.947} \\
 &
  25 &
  29.29/0.828 &
  27.51/0.812 &
  26.93/0.796 &
  29.01/0.815 &
  28.57/0.806 &
  \cellcolor[HTML]{EFEFEF}\underline{30.00/0.854} &
  \cellcolor[HTML]{EFEFEF}\textbf{30.58/0.865} \\
\multirow{-3}{*}{CBSD~\cite{martin2001cbsd}} &
  50 &
  23.06/0.540 &
  25.74/\underline{0.700} &
  24.78/0.695 &
  25.37/0.657 &
  24.75/0.669 &
  \cellcolor[HTML]{EFEFEF}\textbf{27.05/0.712} &
  \cellcolor[HTML]{EFEFEF}\underline{26.85/0.703} \\ \midrule
\multicolumn{2}{c}{Avg. Infer. time (s)} &
  451.9 &
  153.9 &
  147.9 &
  \underline{16.8} &
  149.2 &
  \cellcolor[HTML]{EFEFEF}\textbf{10.1} &
  \cellcolor[HTML]{EFEFEF}51.6 \\ \bottomrule
\end{tabular}%
}}
\vspace{-0.4cm}
\end{table*}
\begin{equation}
\label{equ_ensemble_optim}
\mathop{\arg\min}\limits_{\theta_t} \left\|\hat{M_t}\odot y_t-\hat{M_t}\odot x\right\|_2.
\end{equation}

The optimization task, represented by $L_{rec}(y_t,x)$, learns to reconstruct noisy image cropped by random masks, aligns with pre-training. The alignment minimizes the gap between pre-training and zero-shot inference to avoid over-fitting, and reduces the inference steps required, thus accelerating the denoising process.
Thanks to the well-crafted mechanism, we can accomplish high-quality results with preserved details in reduced time \textbf{without any other regularization}.

\noindent\textbf{Pixel-based iterative refinement.}
For a lower mask ratio and reconstruction of more detailed images, we abandon constraints on unmasked regions in previous optimization goals (Eq.~\ref{equ_pretrain_optim} and Eq.~\ref{equ_ensemble_optim}), thus making information under these areas unreliable, we preserve only the results corresponding to $\hat{M}$ for final denoised outcome $\overline{y}$. However, as one forward pass provide partial denoising results, an ensemble process is crucial. Specially, we employ an Exponential Moving Average (EMA) strategy to optimize the use predictions during iterations with little increase in inference time (Sec.~\ref{sec:ablation}):
\begin{equation}
\label{equ_ensemble_ema}
\overline{y}=\hat{M_t} \odot
(\beta\cdot \overline{y}+(1-\beta)\cdot y_t)+M_t\odot \overline{y}.
\end{equation}

For an in-depth look at proposed ensemble algorithm, see Alg.~\ref{alg1}.
During inference, the pre-trained weights not only provide a better startup but also act as regularization for the network, preventing from over-fitting too early and leading to better performance with less inference time (Sec.~\ref{sec:ablation}).

\subsection{Adaptation to Real-world Noise Removal}
\label{sec:method_realnoise}

Real-world noise exhibit strong spatial correlations, \ie the noise is correlated across adjacent pixels. In such scenarios, employing a straightforward masking mechanism still allows the model to learn information related to noise patterns. To address this problem, we apply larger masking ratios than that used for synthetic noise. Additionally, we integrate a simple Pixel-shuffle Down-sampling (PD) mechanism during zero-shot inference to reduce spatial correlation in noise.

Specifically, instead of directly processing noisy image $x\in \mathbb R^{1\times C\times H\times W}$ in Eq.~\ref{equ_ensemble_recon}, we handle its down-sampled versions $Down(x)\in \mathbb R^{d^2\times C\times \frac{H}{d}\times \frac{W}{d}}$ using simple Pixel-shuffle with factor $d$, and $d^2$ sub-samples are concatenated along batch dimension for joint denoising. Following the same iterative filling mechanism described above, we apply pixel unshuffle to the denoised result $\overline{y}$ to obtain final denoised outcome $Up(\overline{y})$.
We add minimal PD operations to address spatial correlated noise, illustrating the effect of pre-trained weights, performance on real-world noisy dataset can be improved by applying better sub-sampling approaches~\cite{zhou2020pd_denoising,lee2022ap-bsn,pan2023sdap} (Sec.~\ref{sec:discussion}) as they have been intensively studied.
\begin{figure*}[tbp]
\setlength{\abovecaptionskip}{0.1cm}
\setlength{\belowcaptionskip}{0.1cm}
  \centering
   \includegraphics[width=1\linewidth]{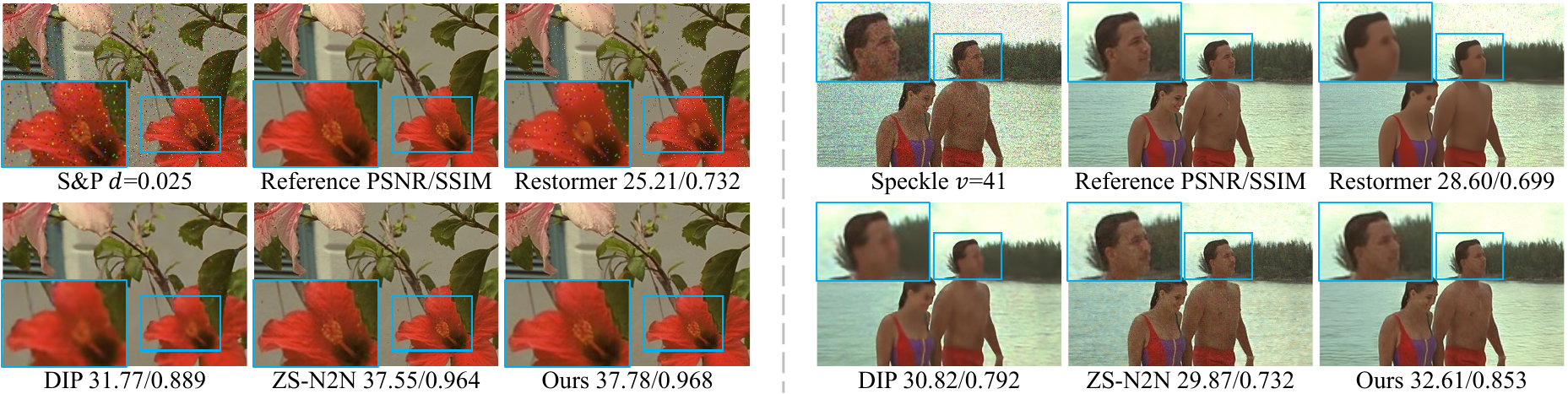}
   \caption{Qualitative results on unseen noise types. Restormer is trained with Gaussian $\sigma$=25. Noisy patches are from kodim07 and kodim12.}
   \vspace{-0.5cm}
   \label{fig_oodnoise}
\end{figure*}
\begin{table*}[t]
\centering
\setlength{\abovecaptionskip}{0.1cm}
\setlength{\belowcaptionskip}{0.1cm}
\caption{Quantitative \textbf{generalization evaluation} results on Kodak. All supervised/unsupervised methods trained on $\sigma$=25 Gaussian, tested on 5 unseen noise types. (Average from all 6 settings.)}
\label{table_general_noise}
\resizebox{1\columnwidth}{!}{%
\begin{tabular}{l|cc|cc|cccc}
\toprule
 &
  \multicolumn{2}{c|}{Supervised} &
  \multicolumn{2}{c|}{Unsupervised} &
  \multicolumn{4}{c}{Zero-shot} \\ \cmidrule(l){2-9} 
\multirow{-2}{*}{Test Noise} &
  SwinIR~\cite{liang2021swinir} &
  Restormer~\cite{zamir2022restormer} &
  Nb2Nb~\cite{huang2021neighbor2neighbor} &
  B2U~\cite{wang2022blind2unblind} &
  DIP~\cite{Ulyanov2018dip} &
  ZS-N2N~\cite{Mansour2023zs-n2n} &
  Ours (faster) &
  Ours \\ \midrule
Gaussian $\sigma$=25 &
  \underline{32.89/0.895} &
  \textbf{33.04/0.897} &
  32.06/0.880 &
  32.26/0.880 &
  30.05/0.806 &
  29.46/0.775 &
  \cellcolor[HTML]{EFEFEF}30.94/0.848 &
  \cellcolor[HTML]{EFEFEF}31.78/0.865 \\
Gaussian $\sigma$$\in${[}10,50{]} &
  27.29/0.628 &
  30.00/0.729 &
  28.68/0.713 &
  29.24/0.726 &
  29.56/0.783 &
  29.36/0.753 &
  \cellcolor[HTML]{EFEFEF}\underline{30.89/0.837} &
  \cellcolor[HTML]{EFEFEF}\textbf{31.66/0.846} \\
Poisson $\lambda$$\in${[}10,50{]} &
  25.06/0.622 &
  26.52/0.683 &
  27.31/0.703 &
  28.22/0.718 &
  28.67/0.758 &
  28.17/0.732 &
  \cellcolor[HTML]{EFEFEF}\underline{29.94/0.826} &
  \cellcolor[HTML]{EFEFEF}\textbf{30.57/0.832} \\
NLF from~\cite{plotz2017darmstadt} &
  \underline{32.52}/0.862 &
  31.71/0.857 &
  31.88/0.859 &
  31.98/0.859 &
  29.71/0.821 &
  31.02/0.834 &
  \cellcolor[HTML]{EFEFEF}32.26/\underline{0.886} &
  \cellcolor[HTML]{EFEFEF}\textbf{33.15/0.901} \\
Speckle $v$$\in${[}10,50{]} &
  31.97/0.841 &
  33.52/0.884 &
  31.31/0.837 &
  31.65/0.847 &
  30.73/0.818 &
  33.78/0.891 &
  \cellcolor[HTML]{EFEFEF}\underline{34.79/0.924} &
  \cellcolor[HTML]{EFEFEF}\textbf{35.79/0.933} \\
S\&P $d$$\in${[}0.02,0.05{]} &
  23.96/0.614 &
  23.63/0.613 &
  27.04/0.686 &
  29.44/0.796 &
  29.54/0.800 &
  \underline{35.25}/0.952 &
  \cellcolor[HTML]{EFEFEF}35.05/\underline{0.953} &
  \cellcolor[HTML]{EFEFEF}\textbf{36.87/0.964} \\ \midrule
Average &
  28.94/0.744 &
  29.73/0.777 &
  29.71/0.800 &
  30.47/0.804 &
  29.71/0.798 &
  31.17/0.823 &
  \cellcolor[HTML]{EFEFEF}\underline{32.31/0.879} &
  \cellcolor[HTML]{EFEFEF}\textbf{33.30/0.890} \\ \bottomrule
\end{tabular}%
}
\vspace{-0.3cm}
\end{table*}

\section{Experiments}
\label{Experiments}
We assess our method against typical methods including DIP~\cite{Ulyanov2018dip}, Noise2Void (N2V)~\cite{krull2019noise2void}, Noise2Self (N2S)~\cite{batson2019noise2self}, Zero-Shot Noise2Noise (ZS-N2N)~\cite{Mansour2023zs-n2n}, and FasterDIP~\cite{liu2023fasterdip}. We modify N2V and N2S to single-image version (N2V* and N2S*). EMA ensemble result of DIP and FasterDIP are reported with their official code. Refer to supplementary material (Supp.) for EMA results of N2V* and N2S*, and comparison with more DIP-based~\cite{jo2021rethinkingdip}, diffusion-based~\cite{wang2022ddnm,garber2024ddpg}, zero-shot modifications from unsupervised methods~\cite{lee2022ap-bsn,zhang2023mm-bsn,jang2024puca}. Only non-ensemble ZS-N2N is presented due to negligible performance differences with EMA version.
We compare Peak Signal-to-Noise Ratio (PSNR) and Structure Similarity Index Measure (SSIM) on synthetic (Sec.~\ref{sec:awgn&poisson}, Sec.~\ref{sec:syn_general}) and real noise (Sec.~\ref{sec:real_noise}). Additional tests on medical images (Sec.~\ref{sec:medical_noise}) show our method's adaptability beyond natural images.

\vspace{-0.4cm}
\subsection{Experimental Setup}
\noindent\textbf{Pre-training.}
Pre-training is performed on two Nvidia RTX 3090 GPUs using Adam optimizer with $\beta_1$=0.9 and $\beta_2$=0.9. Initial learning rate is $2e^{-3}$ and decays to $1e^{-5}$ with cosine annealing strategy over 80K iterations with a batch size of 64.
We initiate pre-training on randomly cropped 256$\times$256 patches from subset of ImageNet~\cite{deng2009imagenet} with around 48,000 images. Two sets of pre-trained weights with masking probability $p$ ($p$=0.3 for synthetic noise and a higher ratio of 0.8$\sim$0.95 for spatially correlated noise) are trained. Further discussion of $p$ is in Sec.~\ref{sec:ablation}.

\noindent\textbf{Zero-shot inference.}
We set learning rate during inference to $2e^{-3}$, and same masking ratio $p$ as pre-training (0.3 for synthetic, 0.8$\sim$0.95 for real noise) is set. EMA weight $\beta$=0.99 for 1000 iterations (specially, 800 iterations for SIDD).
Additionally, with $\beta$=0.9, we achieve performance surpassing most zero-shot methods within 200 iterations, denoted as \textbf{``faster''}. See Supp. for detailed setting.

\subsection{Gaussian \& Poisson Noise}
\label{sec:awgn&poisson}
We investigate Gaussian Noise with $\sigma$$\in$[10,25,50] and Poisson noise with $\lambda$$\in$[10, 25, 50] separately on three datasets: CSet~\cite{dabov2007bm3d}, McMaster~\cite{zhang2011mcmaster} and CBSD~\cite{martin2001cbsd}, with 9, 18 and 68 high-quality images, respectively.
Results are shown in Table~\ref{table_gauss}.
The model is tested across various noise types with same experimental setups, without prior knowledge of noise distribution or intensity.

\noindent\textbf{Analysis.}
DIP tends to produce over-blurry results and struggles especially with intense noise. While ZS-N2N manages to remove weak noise, its simple down-sample approach falters with stronger noise and cause artifacts. As Fig.~\ref{fig_gauss&poiss} illustrates, under Gaussian noise $\sigma$=25 and Poisson noise $\lambda$=25, our method excels in both noise reduction and detail preservation. In some cases, we see an improvement of over 1dB, highlighting the effectiveness of our zero-shot paradigm.

Average inference time is listed in Table~\ref{table_gauss}. Our ``faster'' version achieve the fastest inference speed while surpassing comparing methods in most cases. Even with $\beta=0.99$, our method exhibits competitive inference time and significantly better performance. Params and FLOPs are in Supp.

\begin{figure*}[tbp]
\setlength{\abovecaptionskip}{0.1cm}
\setlength{\belowcaptionskip}{0.1cm}
  \centering
   \includegraphics[width=1\linewidth]{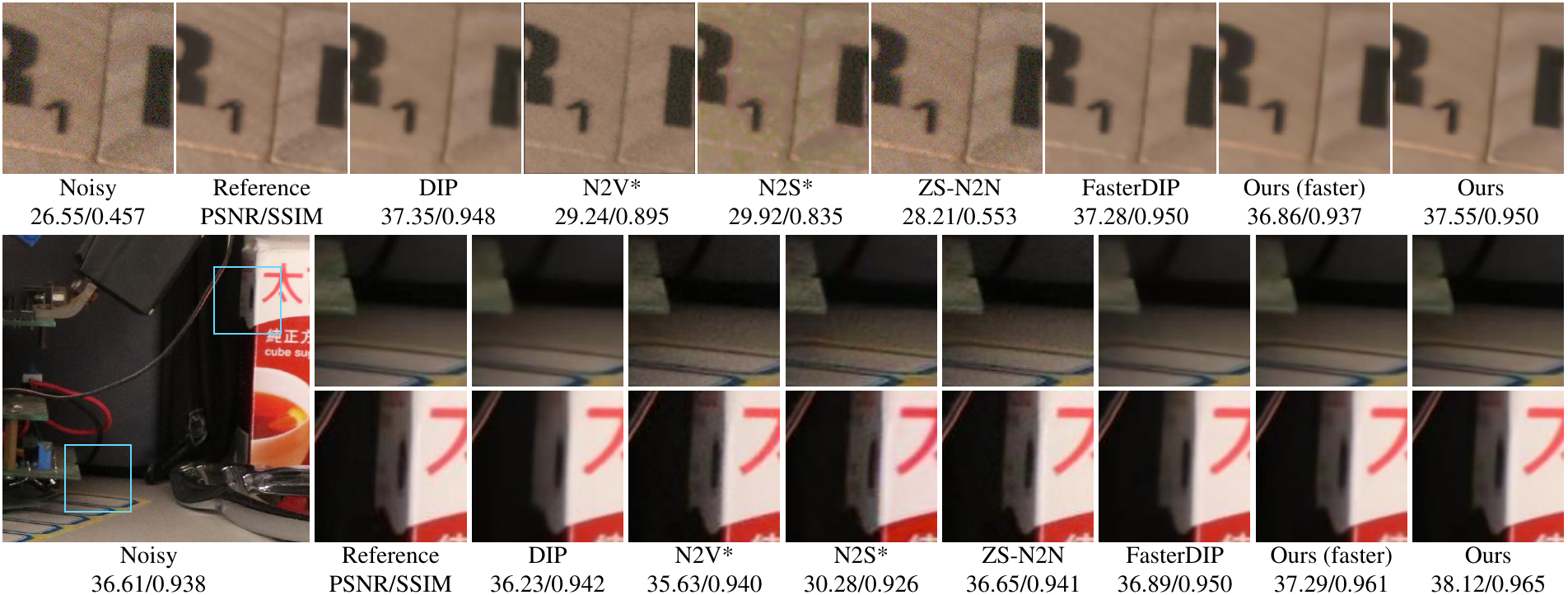}
   \caption{Qualitative results on real noise removal from SIDD and PolyU. Noisy patches are from SIDDval31\_1 and Canon80D\_8\_8\_3200\_ball\_16.}
   \vspace{-0.5cm}
   \label{fig_realnoise}
\end{figure*}
\begin{table}[t]
\centering
\setlength{\abovecaptionskip}{0.1cm}
\setlength{\belowcaptionskip}{0.1cm}
\caption{Quantitative comparison on SIDD, PolyU and FMD for \textbf{real noise removal}.}
\label{table_realnoise}
\setlength{\tabcolsep}{2.5mm}{
\resizebox{0.8\columnwidth}{!}{%
\begin{tabular}{lcccc|c}
\toprule
\multirow{2}{*}{Methods} & \multicolumn{2}{c}{SIDD~\cite{abdelhamed2018sidd}} & \multirow{2}{*}{PolyU~\cite{xu2018polyu}} & \multirow{2}{*}{FMD~\cite{zhang2019fmd}} & \multirow{2}{*}{\begin{tabular}[c]{@{}c@{}}Avg. Infer.\\ time (s)\end{tabular}} \\
\cline{2-3} & \raisebox{-0.3ex}{validation} & \raisebox{-0.5ex}{benchmark} &  &  \\
\midrule
DIP~\cite{Ulyanov2018dip}           & 33.68/0.802          & \underline{33.67}/0.863 & 37.91/0.952          & \underline{32.85}/0.840      & 333.2    \\
N2V*~\cite{krull2019noise2void}           & 26.74/0.627          & 25.34/0.595 & 35.04/0.921          & 29.79/0.817  & 98.1        \\
N2S*~\cite{batson2019noise2self}           & 26.78/0.573          & 26.93/0.658 & 32.82/0.930          & 31.61/0.759   & 114.4       \\
ZS-N2N~\cite{Mansour2023zs-n2n}        & 25.59/0.422          & 25.61/0.559 & 36.04/0.915          & 31.65/0.768   & \underline{15.1}       \\
FasterDIP~\cite{liu2023fasterdip}     & 33.55/0.795          & 33.55/0.859 & \underline{37.99/0.957}          & 32.07/0.821   & 138.2       \\
\rowcolor[HTML]{EFEFEF} 
Ours (faster)   & \underline{33.68/0.828}          & 33.60/\underline{0.896} & 37.62/\underline{0.957}          & 32.68/\underline{0.846}    & \textbf{7.9}      \\
\rowcolor[HTML]{EFEFEF} 
Ours          & \textbf{34.43/0.844} &  \textbf{34.32/0.903} & \textbf{38.11/0.962} & \textbf{32.97/0.847}  & 37.2  \\ \bottomrule  
\end{tabular}%
}}
\vspace{-0.3cm}
\end{table}

\subsection{Generalization on Unseen Noise}
\label{sec:syn_general}
We believe zero-shot denoising with natural image knowledge offers new perspectives on improving generalizability of denoising methods. We select several recent supervised (SwinIR~\cite{liang2021swinir}, Restormer~\cite{zamir2022restormer}) and unsupervised (Neighbor2Neighbor~\cite{huang2021neighbor2neighbor}, Blind2Unblind~\cite{wang2022blind2unblind}) methods trained on Gaussian with $\sigma$=25 for demostration. Testing them on 5 unknown noise types on Kodak~\cite{franzen1999kodak}.

\noindent\textbf{Analysis.}
As illustrated in Table~\ref{table_general_noise} and Fig.~\ref{fig_oodnoise}, although methods trained on multiple noisy images achieve better results on noisy cases with the same distribution, they exhibit poor generalization performance. In contrast, zero-shot methods often perform better generalization capabilities, especially our method which achieves the best performance across all types of generalization noise.

\subsection{Real Noisy Datasets}
\label{sec:real_noise}
We assess denoising capability of MPI on synthetic noise in previous experiments. However, real-world noise is more complicated and challenging. We test on SIDD~\cite{abdelhamed2018sidd} and PolyU~\cite{xu2018polyu} datasets, including 1280 patches from the SIDD validation and 1280 from SIDD benchmark, and all 100 official patches from PolyU to show our paradigm on real images. 
Due to the differences between synthetic and real noise, we report results from comparison methods from their optimal iteration.

\begin{figure}[tbp]
\begin{minipage}[t]{0.45\columnwidth}
\vspace{0cm}
\setlength{\abovecaptionskip}{0.1cm}
\setlength{\belowcaptionskip}{0.1cm}
  \centering
   \includegraphics[width=1\linewidth]{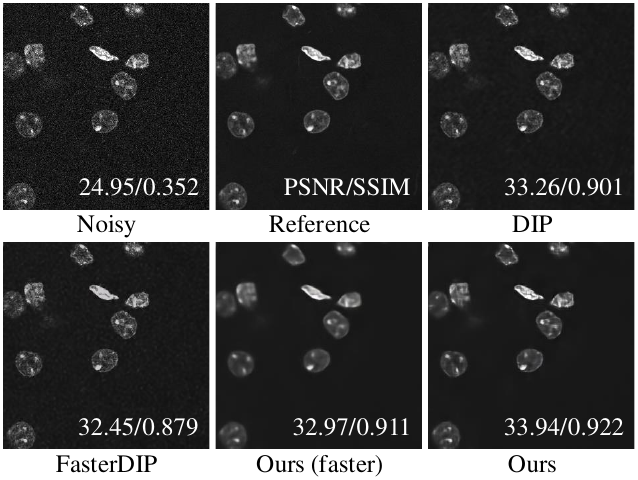}
   \caption{Validation of pre-trained representations on image content differs from natural images. Comparing between ``Baseline'' (w/o pre-train) and ``Ours'' (w pre-train). Noisy patch is from TwoPhoton\_MICE\_3. See quantitative comparison at Table~\ref{table_ab_pretrain}.}
   \vspace{-0.5cm}
   \label{fig_medical}
\end{minipage}
\hfill
\begin{minipage}[t]{0.52\columnwidth}
\vspace{0mm}
\setlength{\abovecaptionskip}{0.2cm}
\setlength{\belowcaptionskip}{0.2cm}
\centering
    \includegraphics[width=1\linewidth]{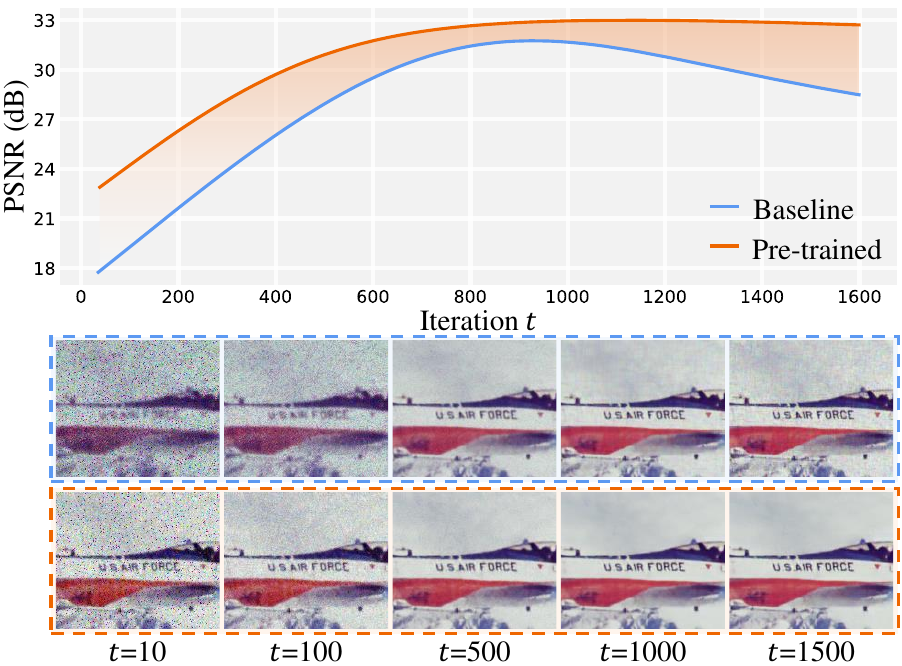}
    \captionof{figure}{Effect of pre-trained model. Examples using Gauss $\sigma$=25 removal on F\_16 with $\beta$=0.99. Pre-trained results are labeled in \textcolor[HTML]{EE6600}{orange}, while default initialized results are labeled in \textcolor[HTML]{5C99F2}{blue}.}
    \label{fig_ab_pretrain}
\end{minipage}
\end{figure}

\noindent\textbf{Analysis.}
As shown in Table~\ref{table_realnoise}, our method excels over other zero-shot approaches on both datasets. This underlines our method's effectiveness on real-world noise removal.
Fig.~\ref{fig_realnoise} show our method's capability to balance noise removal and detail retention. In essence, our method is adept at real-world denoising, offering a robust solution for image quality enhancement in challenging situations.

\subsection{Generalization to Medical Images}
The pre-trained model, which has learned the feature distributions of natural images, raises a question: Can this knowledge be applied to other image types? To answer this question, we select a fluorescence microscopy dataset (FMD)~\cite{zhang2019fmd} characterized by colors and textures distinctly different from natural images, using all released 48 images in testset for evaluation. See Supp. for more image types.

\noindent\textbf{Analysis.}
Our method still excels in denoising performance, as seen in Table~\ref{table_realnoise}. Despite such monochromatic microscopic images are not included in pre-training dataset and exhibits large differences between natural images, pre-trained knowledge still enhances zero-shot denoising performance, as evidenced in Fig.~\ref{fig_medical} and Table~\ref{table_ab_pretrain}, demonstrating the generalizability of pre-trained weights.
\label{sec:medical_noise}

\section{Ablation Study \& Discussion}
\label{sec:ablation}
\subsection{Ablation}
\begin{figure}[t]
\centering
\begin{minipage}[t]{0.5\columnwidth}
\vspace{-2mm}
\hfill
\vspace{-1mm}
\setlength{\abovecaptionskip}{0.1cm}
\setlength{\belowcaptionskip}{0.1cm}
    \centering
   \includegraphics[width=1\linewidth]{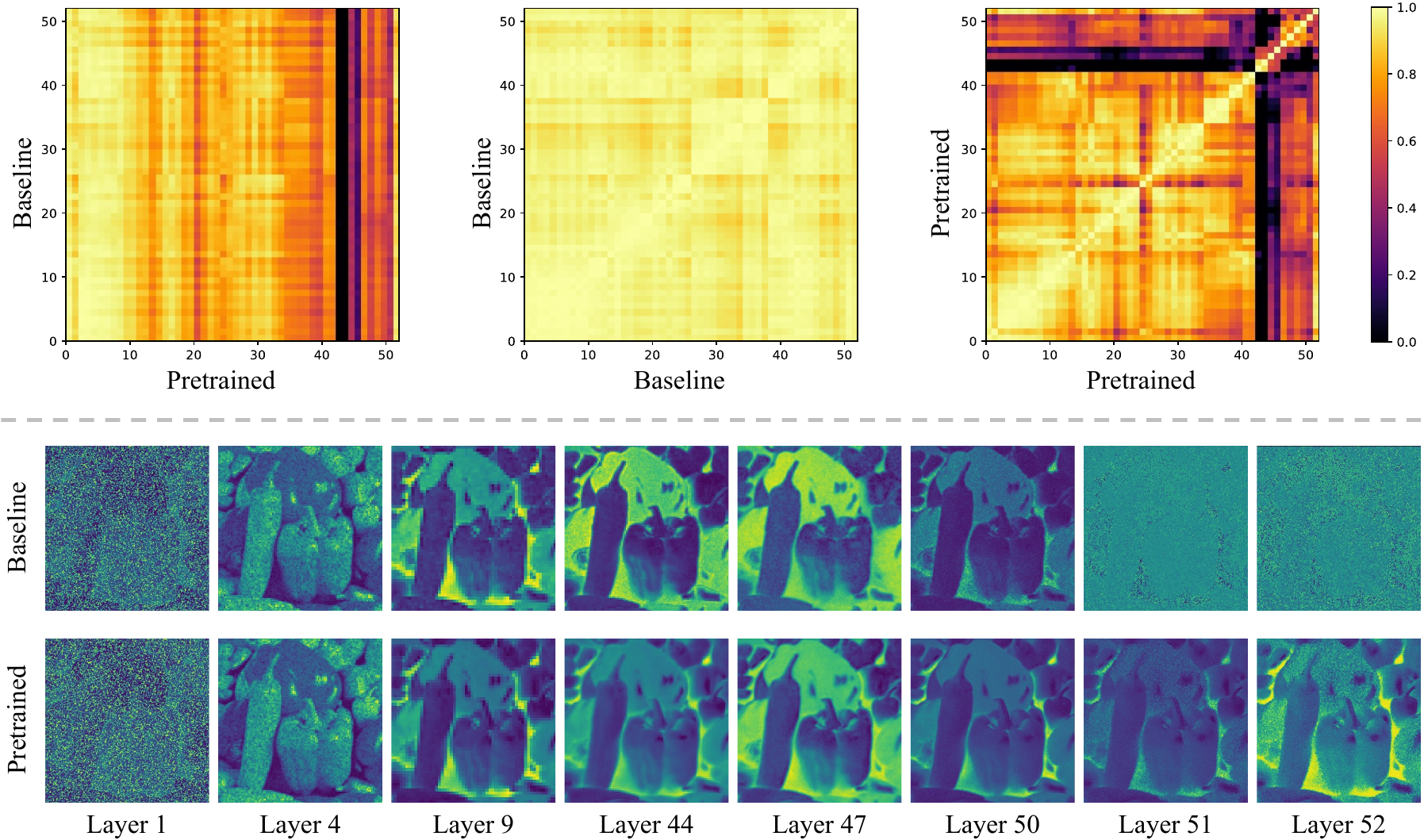}
   \caption{CKA (above)~\cite{kornblith2019cka} and PCA (below) visualization of features extracted from the final timestep of model. Distribution of pre-trained model (``Pretrained'') and from scratch (``Baseline'') during inference is significantly different in last layers. Pre-trained model tends to restore the complete image, while the baseline model primarily focusing on restoring masked regions only.}
   \label{cka_feature}
\end{minipage}%
\hfill
\begin{minipage}[t]{0.45\columnwidth}
\vspace{0mm}
\setlength{\abovecaptionskip}{0.04cm}
\setlength{\belowcaptionskip}{0.04cm}
\centering
    \includegraphics[width=1\linewidth]{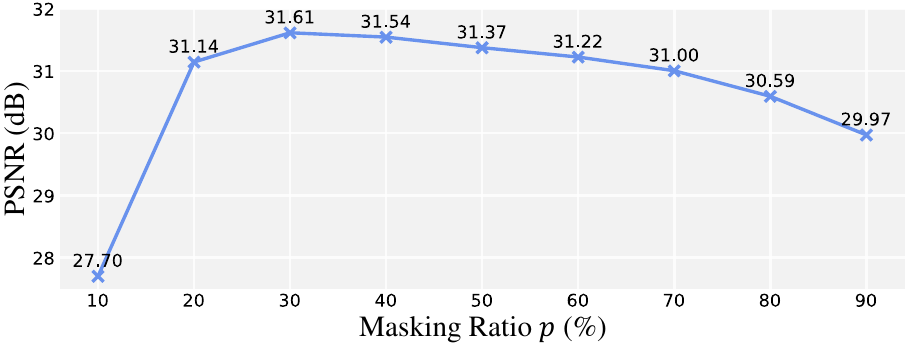}
    \captionof{figure}{Effect of masking ratios. 30\% balances noise removal and prevents over-smoothing for synthetic noise.}
    \label{fig_ab_maskratio}
    \vspace{2pt} 
\centering
\setlength{\abovecaptionskip}{0.04cm}
\setlength{\belowcaptionskip}{0.04cm}
\captionof{table}{Ablation of pre-training, with default settings noted in \colorbox[HTML]{EFEFEF}{gray}. ``SIDD'' denotes SIDD validation.}
\label{table_ab_pretrain}
\setlength{\tabcolsep}{1.3mm}{
\resizebox{\columnwidth}{!}{%
\begin{tabular}{ccccc}
\toprule
$\beta$                & Pretrain                  & CSet~\cite{dabov2007bm3d}                                & SIDD~\cite{abdelhamed2018sidd}                                & FMD~\cite{zhang2019fmd}                                 \\ \midrule
                       & \cellcolor[HTML]{EFEFEF}\checkmark & \cellcolor[HTML]{EFEFEF}31.61/0.841 & \cellcolor[HTML]{EFEFEF}34.43/0.844 & \cellcolor[HTML]{EFEFEF}32.97/0.847 \\
\multirow{-2}{*}{0.99} & \ding{55}                         & 30.90/0.811                         & 32.31/0.746                         & 31.44/0.786                         \\ \midrule
                       & \cellcolor[HTML]{EFEFEF}\checkmark & \cellcolor[HTML]{EFEFEF}30.83/0.824 & \cellcolor[HTML]{EFEFEF}33.68/0.828 & \cellcolor[HTML]{EFEFEF}32.68/0.846 \\
\multirow{-2}{*}{0.90} & \ding{55}                         & 30.10/0.806                         & 33.42/0.824                         & 32.31/0.833                         \\ \bottomrule
\end{tabular}%
}}
\end{minipage}
\vspace{-0.4cm}
\end{figure}

\noindent\textbf{Pre-trained weights.}
Building on Sec.~\ref{sec:motivation}, we question the role of pre-trained weights in zero-shot inference by comparing inference with pre-trained weights and optimizing from scratch, resembling a standard blind-spot network~\cite{krull2019noise2void}. As depicted in Fig.~\ref{fig_ab_pretrain} and Table~\ref{table_ab_pretrain}, the latter quickly peaks and risks over-fitting due to the simple task of content recovery from masked images, making it challenging to specify an optimal iteration for all images. Conversely, the pre-trained model achieves better initial performance and maintains close-to-optimal performance for more extended period of time.

Unlike other zero-shot techniques train models from scratch for a single image to learn noise-resistant image content, we offer a new perspective by showing that a pre-trained model can aid in zero-shot tasks. The pre-trained weights, encapsulating views from multiple natural images, making it more robust to iteration count and provides better options for faster zero-shot denoising.

Moreover, We investigated the impact of pre-training on inference at the hidden layer level (see Fig.~\ref{cka_feature}). Features extracted with pre-trained weights exhibit significant divergence from those produced by the baseline, \ie, the usual zero-shot denoising approach. Specifically, pre-trained model restores the complete image, with more distinct features between layers, whereas the baseline model's features are less differentiated between layers, tending to only restore the masked parts, which may result in sub-optimal convergence towards local minima.

\noindent\textbf{Masking ratios.}
Fig.~\ref{fig_ab_maskratio} shows impact of different masking ratios on denoising with Gaussian $\sigma$=25. 
Lower masking ratios fails to completely remove noise, while higher masking ratios can cause overly smoothed results. A 30\% masking ratio balances detail preservation and noise reduction for synthetic noise. However, a higher $p$ of 0.8$\sim$0.95 is required for real-world noise. See Supp. for more details.

\noindent\textbf{Ensemble strategy.}
We explore ensemble strategies, including EMA-based (``EMA''), straightforward averaging during iterations (``Average'') and average after specific optimization step (``Avg after 500e'', where 500 is optimal).
Due to the inability to obtain predictions for all pixels in a single forward pass (Sec.~\ref{sec:zeroshot_denoise}), the ``w/o Ensemble'' result comes from the final prediction for each pixel. And ``Last'' provided for results of final forward prediction, a significant performance drop is caused by unreliable pixels (For details, see Supp.). 
Additionally, to validate our mask-based ensemble strategy, we remove masks from Eq.~\ref{equ_ensemble_optim} and Eq.~\ref{equ_ensemble_ema}, with full-pixel loss and ensemble (``EMA w/o Mask'').
See results in Table~\ref{table_ensemblemode}. Proposed ``EMA'' achieve significant better performance, aiding denoising with efficiency.
\noindent
\begin{table}[t]
\vspace{-0.3cm}
\centering
\noindent
\begin{minipage}[t]{0.5\columnwidth}
\vspace{2mm}
\centering
\centering
\setlength{\abovecaptionskip}{0.1cm}
\setlength{\belowcaptionskip}{0.1cm}
\captionof{table}{Ablation of ensemble strategy. ``Time (s)'' denotes Infer. time (s)}
\label{table_ensemblemode}
\setlength{\tabcolsep}{1mm}{
\resizebox{0.9\columnwidth}{!}{%
\begin{tabular}{lcc}
\toprule
Ensemble strategy       & PSNR/SSIM   & Time (s) \\ \midrule
Avg after 500e & 30.88/0.793 & 48.3           \\
Average             & 31.28/0.835 & 49.7            \\
EMA w/o mask        & 23.48/0.441 & 52.1            \\
w/o Ensemble        & 30.23/0.797 & 51.7           \\
Last                & 13.73/0.154 & 49.0           \\
\rowcolor[HTML]{EFEFEF} 
EMA                 & 31.61/0.841 & 53.5           \\
\bottomrule
\end{tabular}%
}}
\end{minipage}%
\hfill
\begin{minipage}[t]{0.45\columnwidth}
\centering

\centering
\setlength{\abovecaptionskip}{0.0cm}
\setlength{\belowcaptionskip}{0.0cm}
\captionof{table}{Discussion on over-fitting.
}
\label{table_discuss}
\setlength{\tabcolsep}{1.3mm}{
\resizebox{\columnwidth}{!}{%
\begin{tabular}{lcccc}
\toprule
\multirow{2}{*}{\begin{tabular}[c]{@{}c@{}}Methods/\\ PSNR\end{tabular}} & \multicolumn{3}{c}{Iteration} & \multirow{2}{*}{\begin{tabular}[c]{@{}c@{}}Avg. Infer.\\ time (s)\end{tabular}} \\ 
\cline{2-4} & \raisebox{-0.3ex}{1,000} & \raisebox{-0.5ex}{1,100}  & \raisebox{-0.5ex}{1,500}      &    \\ 
\midrule
Ours    & 31.61 & 31.58                    & 31.35 & 62.6 \\
Ours+ES~\cite{shi2022measuringdip} & 31.66 & 31.65                    & 31.66 & 51.7 \\ \bottomrule
\end{tabular}%
}}

    \vspace{0.2cm}
\centering
\setlength{\abovecaptionskip}{0.0cm}
\setlength{\belowcaptionskip}{0.0cm}
\captionof{table}{Discussion of downsampling on SIDD validation with 256$\times$256.}
\label{table_discuss_pd}
\setlength{\tabcolsep}{1.5mm}{
\resizebox{0.7\columnwidth}{!}{%
\begin{tabular}{@{}lcc@{}}
\toprule
     & PSNR/SSIM   & Time (s) \\ \midrule
+PD  & 34.42/0.843 & 29.6     \\
+RSG~\cite{pan2023sdap} & 34.75/0.852 & 38.3     \\ \bottomrule
\end{tabular}%
}}
\end{minipage}
\vspace{-2mm}
\end{table}

\subsection{Discussion}
\label{sec:discussion}
Our method uses masking and minimal PD during inference to highlight pre-training's role without explicit regularization. We now explore further enhancements during inference with more strategies.

\noindent\textbf{Over-fitting.}
Although pre-training mitigates over-fitting for synthetic and real noise, the over-parameterized network may still learn noise patterns over time due to lack of explicit mechanisms to avoid identity mapping.
This is common challenge in many zero-shot models, we suggest early-stopping~\cite{shi2022measuringdip} (Table~\ref{table_discuss},``ES'' for early-stopping) to avoid over-fitting and reduce inference time.

Additionally, we compare with other over-fitting prevention methods, \eg, TV regularization of output and augmentation to input image. These approaches either resulted in suboptimal performance or longer inference times. More details on over-fitting and prevention strategies can be found in Supp.

\noindent\textbf{Sub-sampling.} Shown in Sec.~\ref{sec:method_realnoise}, minimal pixel-shuffle is used to reduce spatial correlation in real-world noise, but may cause chessboard artifact and reduce performance due to its regular down-sampling strategy. Better down-sampling strategies have been widely studied, and here we choose RSG~\cite{pan2023sdap} to illustrate, see results at Table~\ref{table_discuss_pd}. For more comparison and visual comparison, see Supp.
\section{Acknowledgements}
This work was supported in part by the National Natural Science Foundation of China under Grant 62401532,
in part by the Anhui Provincial Key Research and Development Plan 202304a05020072,
and in part by the Anhui Provincial Natural Science Foundation 2308085QF226,
in part by the Fundamental Research Funds for the Central Universities WK2090000065,
and in part by the China Postdoctoral Science Foundation under Grant 2022M720137.

\clearpage
{
\small
\bibliographystyle{unsrt}
\bibliography{main}
}

\clearpage
\setcounter{page}{1}
\setcounter{section}{0}

\renewcommand\thesection{\Alph{section}}
\section{Introduction}
\label{sec:supp_intro}
This document provides supplementary materials for the
main paper. Specifically, a brief conclusion of works related to ours is in Sec.~\ref{sec:related_works} Sec.~\ref{sec:add_details_mpi} presents more details and demonstrations about proposed iterative filling (Sec.~\ref{sec:add_details_iterative_filling}, Sec.~\ref{sec:discard_unreliable_pixels}) and different strategies to adapt to real-world spatial-correlated noise (Sec.~\ref{sec:strategies_realnoise}).
Sec.~\ref{sec:add_discussion} presents more discussion about masking ratio (Sec.~\ref{sec:add_discussion_maskratio}), over-fitting (Sec.~\ref{sec:add_discussion_overfit}) and different downsampling strategies (Sec.~\ref{sec:add_discussion_downsample}).
Additionally, we conduct further analysis of pre-training (Sec.~\ref{sec:add_pretrain}) and expanded our framework to other network structures (Sec.~\ref{sec:add_network}). More experimental details and qualitative comparison results can be found in Sec.~\ref{sec:add_expset&result} and Sec.~\ref{sec:add_qualitative}.
\section{Related Works}
\label{sec:related_works}

\subsection{Unsupervised Image Denoising}
Unlike supervised approaches~\cite{zhang2017dncnn,zhang2018ffdnet,liang2021swinir,zamir2022restormer}, unsupervised denoising focuses on situations when paired data is unavailable. Methods in this category include:

\textbf{Paired noisy-noisy images.} 
To learn consistent representations from varied noise observations of the same scene, Lehtinen \etal~\cite{lehtinen2018noise2noise} train on mapping from two noisy observations of the same scene. Additional approaches utilize synthetic noise to generate noisy pairs, as seen in~\cite{xu2020nac,moran2020noisier2noise,pang2021recorrupted2recorrupted,zhang2022idr,neshatavar2022cvf_sid}, as well as~\cite{chen2023med} learns shared latent from multiple noise observations.

\textbf{Unpaired noisy-clean images.} Du \etal~\cite{du2020invariant} propose to learn decoupled representations of contents and noise from images. Lin \etal~\cite{lin2023self-collaboration} extend this by using separated noise representations to guide noise synthesis, thereby enhancing the denoising process. Additionally, Wu \etal~\cite{wu2020dbsn} employ a distillation loss from both real and synthetic noisy images.

\textbf{Noisy images only.} Techniques like blind-spot~\cite{laine2019high,wang2022blind2unblind,wang2023lg-bpn,xie2020noise2same,krull2019noise2void,batson2019noise2self}, substitution followed by image reconstruction~\cite{krull2019noise2void,batson2019noise2self}, multiple sub-sampled images from a single noisy scene~\cite{huang2021neighbor2neighbor} or above approaches combined~\cite{lee2022ap-bsn,pan2023sdap,jang2023c-bsn} are developed when only one observation is available from a noise scene. 
Li \etal~\cite{li2023spatially} integrate blind spot strategies and structural insights for adaptive denoising. Score matching and posterior inference are also utilized in~\cite{kim2021noise2score,kim2022noise2score_var}.

In the context of zero-shot denoising tasks, only a single noisy image is visible during training, presenting greater challenges than methods described above.

\subsection{Zero-shot Image Denoising}
Compared to unsupervised methods, zero-shot denoising is more challenging as it aims to train a network for denoising when a single noisy image available.
Typical strategies involve utilizing spatial correlations~\cite{dabov2007bm3d,maggioni2012bm4d}, variation based priors~\cite{cheng2023scoreprior,yue2019variational} or low-frequency characteristics of images, corrupting and reconstructing part of images~\cite{krull2019noise2void,batson2019noise2self,quan2020self2self}, or constructing paired training sets from sub-sampled noisy images~\cite{lequyer2021noise2fast,Mansour2023zs-n2n}.
Among which Noise2Void~\cite{krull2019noise2void} is initially designed for learning from multiple noisy images, shows promise in its zero-shot version. Noise2Self~\cite{batson2019noise2self}reconstructs cyclically masked regions of input noisy image. Although there is a gap compared to supervised or unsupervised methods.
While dropout-ensemble~\cite{quan2020self2self} is adapted on a noisy image for better performance, it leads to over-smoothing and incurs large computational costs.
Noise2Fast~\cite{lequyer2021noise2fast} and Zero-Shot Noise2Noise~\cite{Mansour2023zs-n2n} are fast but struggle to completely remove noise from images, especially spatially-correlated real noise, resulting in suboptimal visual results.
DIP~\cite{Ulyanov2018dip} and its variants~\cite{heckel2018deep-decoder,darestani2021convdecoder} exploits the features of deep networks to learn mappings from random noise to images.
Early stopping~\cite{shi2022measuringdip} or other approaches~\cite{jo2021rethinkingdip,arican2022isnasdip} are used to prevent over-fitting, FasterDIP~\cite{liu2023fasterdip} further discusses the influence of network structure on its performance. However, current zero-shot methods often takes a long time, and parameter settings are carefully selected for various image contents and noise degradation.

\subsection{Masked Image Modeling}
Masked Image Modeling (MIM) helps in learning pre-trained representations for downstream tasks by masking a portion of input images~\cite{he2022mae,xie2022simmim,bao2021beit} and training models to predict the masked contents.
Due to its impressive effects in high-level tasks~\cite{mao2023masked_motion_predict,zhai2023masked_class_incremental_learning}, Masked Image Modeling has also found applications in low-level visual tasks. For instance, Wang \etal~\cite{wang2023context-aware} applies random patch masks during the pre-training of image deraining and desnowing models in handling adverse weather conditions. Zheng \etal~\cite{zheng2023empowering-lowlight} integrates Masked Autoencoder (MAE) to learn illumination-related structural information in a supervised low-light enhancement framework. Notably, despite the successful applications of Masked Image Modeling in several low-level vision tasks, its application in the pre-training scheme for denoising models has not yet been explored.

\section{More Details about MPI}
\label{sec:add_details_mpi}
\subsection{More Details about Iterative Filling}
\label{sec:add_details_iterative_filling}
\begin{figure*}[h]
\setlength{\abovecaptionskip}{0.1cm}
\setlength{\belowcaptionskip}{0.1cm}
\begin{center}
\includegraphics[width=0.7\textwidth]{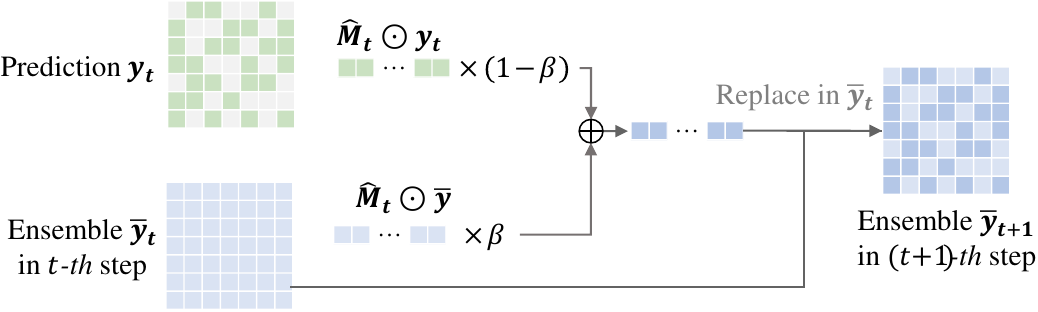}
\end{center}
\caption{
Details of EMA process in Iterative filling. Masked regions of predictions from each optimization step $t$ is assembled to a ensemble $\overline{y}$.
}
\vspace{-0.5cm}
\label{fig_detail_iterative_filling}
\end{figure*}

In the main paper, we mention that Iterative Filling is optimization steps based on pre-trained weights for zero-shot denoising. And as depicted in Sec. \textcolor{red}{3.3} in main paper, to fully leverage the results of each optimization step and preserve more image detail, only the masked regions $\hat{M_t}\odot y_t$ are constrained by the loss at each optimization timestep and considered reliable, others are labeled as ``unreliable pixels'' and discarded (see Sec.~\ref{sec:discard_unreliable_pixels} to know why we need ``unreliable pixels''). During the iterative optimization, we maintain an ensemble version $\overline{y}$, assembling reliable parts $\hat{M_t}\odot y_t$ of each prediction $y_t$ via EMA while keeping the rest unchanged, as shown in Fig.~\ref{fig_detail_iterative_filling}.

With sufficient iterations, ensemble from each pixel is ensemble from hundreds of predictions, ensuring a high-quality ensemble outcome.

\subsection{Why Discard ``Unreliable Pixels''}
\label{sec:discard_unreliable_pixels}
During the zero-shot denoising process, unmasked parts of each optimization result $M_t\odot y_t$ needed be discarded, referred to as ``unreliable pixels'', primarily due to the following reasons:

\noindent\textbf{1)} The pre-training task is set to reconstruct masked regions, that is only masked areas are constrained for reconstruction, while pixels in unmasked areas significantly differ from the actual pixels in the image. This discrepancy might result from the skip connections in the network architecture of DIP~\cite{Ulyanov2018dip}. To maximize the utilize of pre-trained weight and avoid conflicting, pixels corresponding to these unmasked regions should not be considered during optimizing.

\noindent\textbf{2)} For spatially uncorrelated noise, we employ an extremely low mask ratio and distinct mask settings for each color channel to preserve as much image information as possible. In this scenario, it is impractical to expect the network to retain all pixel values in its output, as this can easily lead to identity mapping.

\noindent\textbf{3)} While partial convolution~\cite{liu2018pconv} or others designed for image inpainting can mitigate this problem, they often lead to sub-optimal performance with risks of over-fitting to noise, and require specialized network architectures, limiting the adaptability of proposed framework to other network structures.

For our zero-shot denoising framework, which obtains denoised images via iterative optimization, using a portion of noisy image as cues and training network to complete this "fill-in-the-blank" task proves most effective. Moreover, the optimization generates reconstructions with various masks in iterations, assembling these results to achieve final denoised prediction requires only maintaining an ensemble result additionally, incurring negligible time and space resources.

\subsection{Different Strategies in Dealing with Synthetic \& Real Noise}
\label{sec:strategies_realnoise}
\begin{figure*}[t]
\setlength{\abovecaptionskip}{0.1cm}
\setlength{\belowcaptionskip}{0.1cm}
\begin{center}
\includegraphics[width=0.95\textwidth]{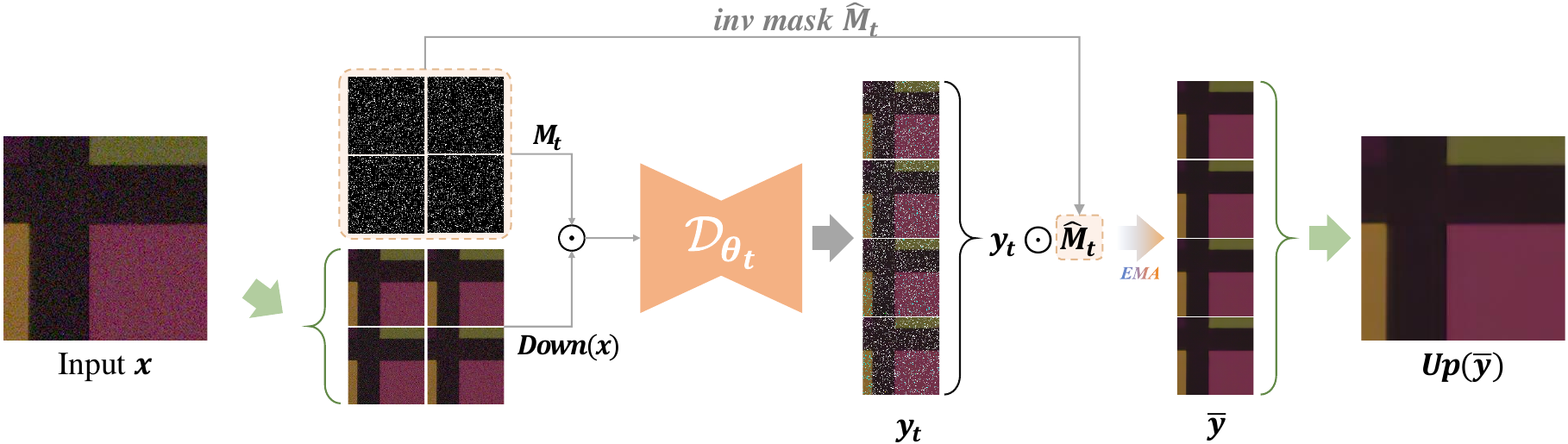}
\end{center}
\caption{
An overview of the zero-shot denoising stage with adaptation to real-world noise. We adapt
downsample ``$Down(\cdot)$'' and upsample ``$Up(\cdot)$'' to achieve noisy subsamples with less spatial correlation in noise, labeled in green arrows, and larger masking ratio is used to further deal with remaining spatial correlations. Actually, not all real noisy images needed to be sub-sampled, we only adapt ``$Down(\cdot)$'' and ``$Up(\cdot)$'' to SIDD dataset.
}
\vspace{-0.5cm}
\label{fig_framework_realnoise}
\end{figure*}

We adapt downsample ``$Down(\cdot)$'' and upsample ``$Up(\cdot)$'' to achieve noisy subsamples with less spatial correlation in noise, and larger masking ratio 80\%$\sim$95\% (90\% for SIDD and 85\% for others) with a unified mask for all channels is used to further deal with remaining spatial correlations. Actually, not all real noisy images needed to be sub-sampled, we only adapt ``$Down(\cdot)$'' and ``$Up(\cdot)$'' to SIDD dataset. See Fig.~\ref{fig_framework_realnoise} for a simple framework, which specifically designed for real-world noise is labeled in green arrows.

\section{Additional Discussion}
\label{sec:add_discussion}
\subsection{Masking Ratio}
\label{sec:add_discussion_maskratio}
In the main paper, we discussed the optimal masking ratio for removing spatially uncorrelated synthetic noise and notes that significantly larger mask ratios are used for real noise.
For real spatially correlated noise, the situation becomes more complex, with no single optimal mask ratio. Based on experience, the most effective mask range is between 80\%$\sim$95\%, which is influenced by the noise's spatial correlation, noise intensity, and the information in image.

\begin{figure}[b]
\setlength{\abovecaptionskip}{0.1cm}
\setlength{\belowcaptionskip}{0.1cm}
  \centering
    \includegraphics[width=0.8\linewidth]{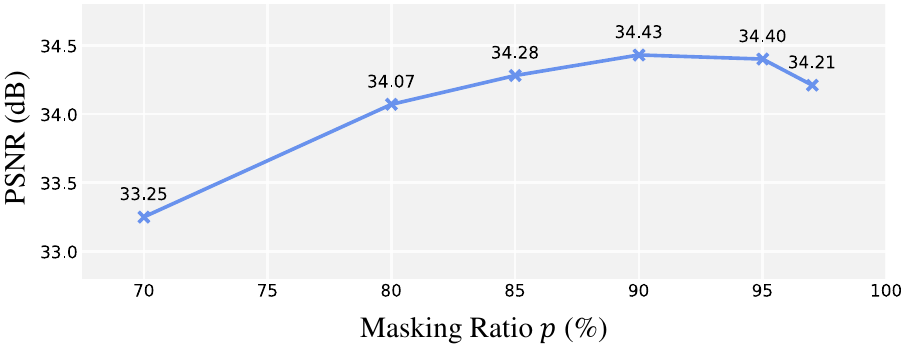}
    \captionof{figure}{Effect of masking ratios on real-world noise (SIDD validation). 90\% for removing strong spatial correlated noise.}
    \label{fig_add_maskratio_sidd}
   \vspace{-0.3cm}
\end{figure}

For the SIDD dataset, we investigated the impact of the masking ratio on SIDD validation, as shown in Fig.~\ref{fig_add_maskratio_sidd}, finding that a 90\% masking ratio is optimal. This is attributed to SIDD images containing limited content information and the noise exhibiting strong spatial correlation.

\begin{figure}[h]
\setlength{\abovecaptionskip}{0.1cm}
\setlength{\belowcaptionskip}{0.1cm}
  \centering
   \includegraphics[width=0.4\linewidth]{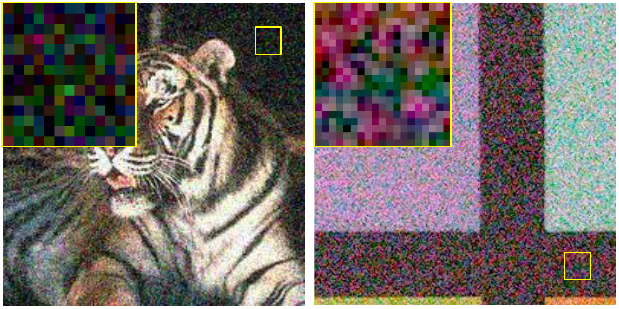}
   \captionof{figure}{Illustration of spatial-correlated real-world noise (right) and synthetic noise (left).}
   \vspace{-0.3cm}
   \label{fig_noise_visual}
\end{figure}

However, the optimal masking ratio for SIDD differs from that for synthetic noise, primarily due to the spatial correlation of the noise within the image. Synthetic noise is spatially uncorrelated, meaning noise signals at neighboring positions do not influence each other. In contrast, real noise, after undergoing a series of ISP processes, exhibits a more complex distribution, resembling blurred spots rather than independent points (see Fig.~\ref{fig_noise_visual}). For synthetic noise, selecting a small masking ratio allows for quicker recovery of image details. Conversely, for real noise, a small masking ratio may lead to the model fitting the noise distribution by relying on neighboring pixel values. In such cases, a larger masking ratio helps mitigate the influence of noise.

\subsection{Over-fitting \& Regularization}
\label{sec:add_discussion_overfit}
\begin{figure}[h]
\vspace{-0.2cm}
\setlength{\abovecaptionskip}{0.1cm}
\setlength{\belowcaptionskip}{0.1cm}
  \centering
    \includegraphics[width=0.6\linewidth]{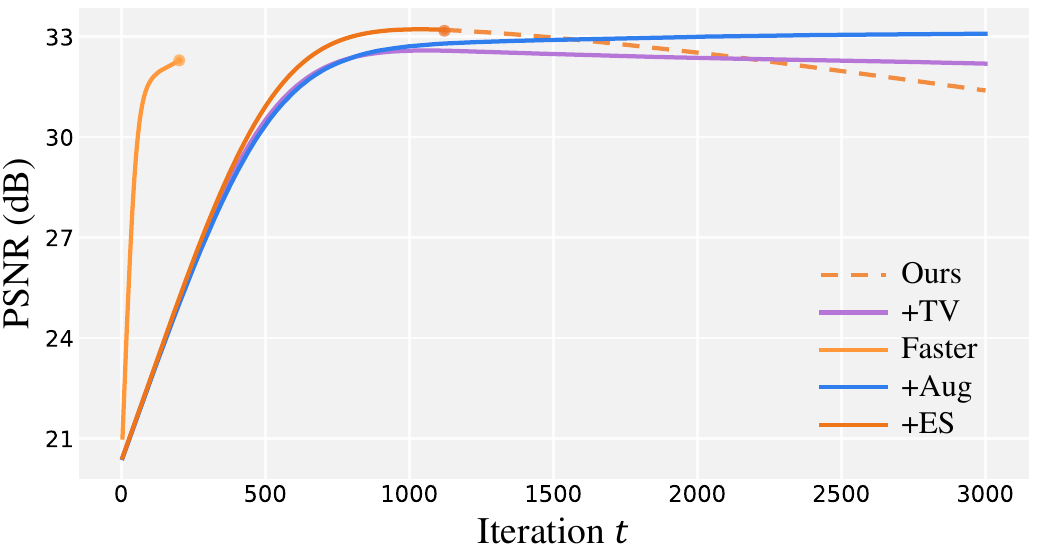}
    \captionof{figure}{Influence of different regularization strategies during iterations, including Total Variation (``+TV''), random image augmentation (``+Aug''), and early stopping (``+ES''). "Ours" and "Faster" are the methods evaluated in mainpaper. Example is tested on F16\_512rgb with Gaussian $\sigma$=25.}
    \label{fig_add_abl_reg}
   \vspace{-0.5cm}
\end{figure}

In the main text, we highlighted that our proposed zero-shot denoising framework still faces over-fitting issues with increased iteration counts, with Table~\ref{table_add_overfit_sidd} and Fig.~\ref{fig_add_abl_reg} showing this problem's impact across more datasets and more regularization. To address this issue, after validation and comparison, we recommend adopting a simple early-stopping strategy to prevent over-fitting—a straightforward, effective approach without additional computational costs. We also compared other strategies:

\begin{itemize}
\item Employing TV regularization helps against overfitting but still leads to performance drop and lower peak PSNR as iterations increase.
\item Adding random transformations to the input image, including flips and random translations, lead to steady performance improvement and higher peak PSNR over more iterations, but increase inference time.
\item Early stopping stops close to peak performance with minimal calculation, providing stable, high-quality results with no added time.
\end{itemize}

\begin{table}[]
\centering
\setlength{\abovecaptionskip}{0.1cm}
\setlength{\belowcaptionskip}{0.1cm}
\caption{Extension of proposed pre-training strategy into other network architectures. A performance improvement can be observed for both settings of beta=0.9 and 0.99 in the experiment across various network architectures.}
    \label{table_add_overfit_sidd}
    \setlength{\tabcolsep}{1.3mm}{
    \resizebox{0.5\columnwidth}{!}{%
    \begin{tabular}{lcccc}
    \toprule
    \multirow{2}{*}{\begin{tabular}[c]{@{}c@{}}Methods/\\ PSNR\end{tabular}} & \multicolumn{3}{c}{Iteration} & \multirow{2}{*}{\begin{tabular}[c]{@{}c@{}}Avg. Infer.\\ time (s)\end{tabular}} \\ 
    \cline{2-4} & \raisebox{-0.5ex}{800} & \raisebox{-0.5ex}{900}  & \raisebox{-0.5ex}{1,200}      &    \\ 
    \midrule
    Ours    & 34.43 & 34.34                    & 33.48 & 39.9 \\
    Ours+ES~\cite{shi2022measuringdip} & 34.64 & 34.63                    & 34.60 & 34.5 \\ \bottomrule
    \end{tabular}%
    }}
\end{table}

\subsection{Down-sampling in Real-world Denoising}
\label{sec:add_discussion_downsample}
\begin{figure}[h]
\setlength{\abovecaptionskip}{0.1cm}
\setlength{\belowcaptionskip}{0.1cm}
  \centering
   \includegraphics[width=0.75\linewidth]{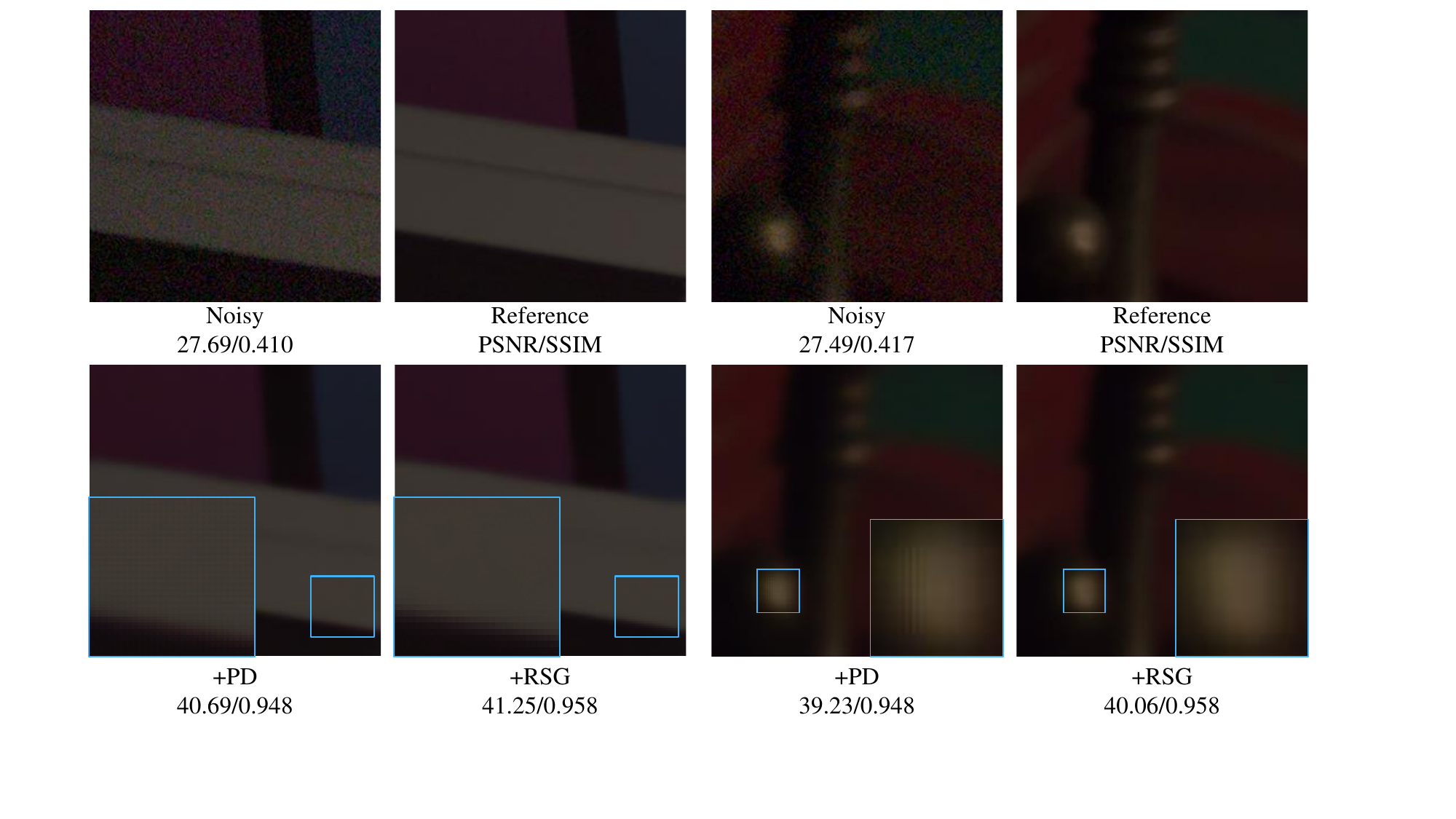}
   \caption{Validation of different down-sampling strategies in real-world denoising. Better down-sampling strategies can further enhance denoising performance of our pipeline. Noisy patches are from SIDDval\_12\_2 and SIDDval\_20\_3.}
   \vspace{-0.5cm}
   \label{fig_pd_vs_rsg}
\end{figure}

As illustrated in Fig.~\ref{fig_framework_realnoise}, specialized downsampling is employed to reduce the spatial correlation of real noise, with different downsampling strategies yielding varying outcomes. Simple pixelshuffle (``+PD'') can easily lead to checkerboard artifacts, whereas more randomized sub-sampling strategies~\cite{pan2023sdap} (``+RSG'') can more effectively disrupt noise spatial correlations and, due to their randomness, avoid checkerboard artifacts, as depicted in Fig.~\ref{fig_pd_vs_rsg}. A potential issue with this approach is the over-smoothed denoising predictions. Therefore, downsampling is only applied in strong spatially-correlated real noise.

\vspace{-0.3cm}
\section{Additional Analysis on Pre-training}
\label{sec:add_pretrain}
\subsection{Noise Intensity}
\begin{figure}[h]
\setlength{\abovecaptionskip}{0.1cm}
\setlength{\belowcaptionskip}{0.1cm}
  \begin{minipage}[b]{1\textwidth} 
    \centering
   \includegraphics[width=0.7\linewidth]{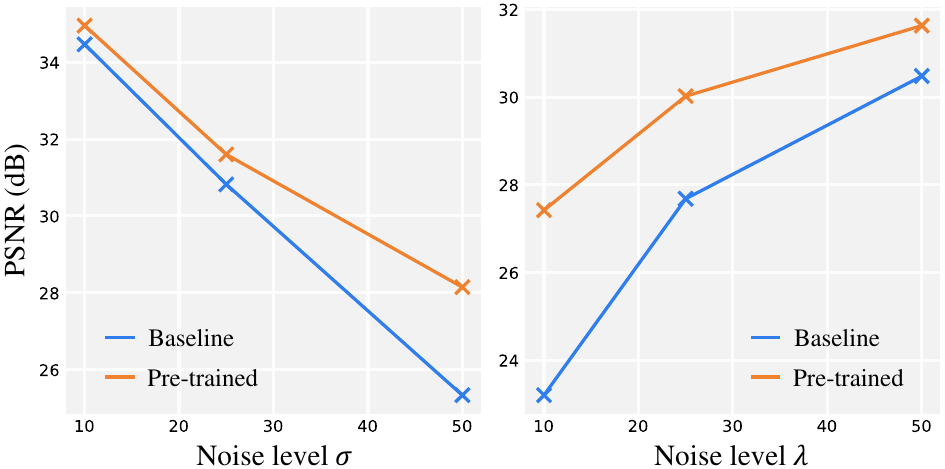}
   \caption{Effect of pre-training on different noise levels from Gaussian (left) and Poisson (right) on CSet. Pre-training is beneficial for all 6 noise levels, especially in cases of intense noise.}
   \label{fig_pretrain_noiselevel}
   \end{minipage}

   \begin{minipage}[b]{1\textwidth} 
  \centering
   \includegraphics[width=0.7\linewidth]{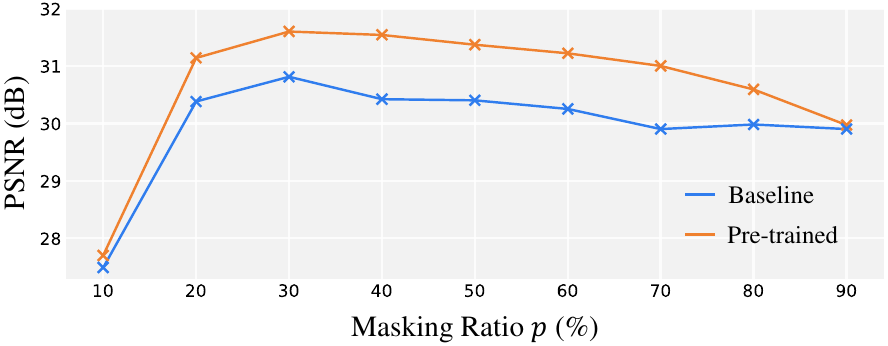}
   \caption{Effect of pre-training on different masking ratios $p$ from Gaussian noise on CSet. Pre-training is beneficial for all masking ratios, especially in cases of $20 \leq p \leq 80$.}
   \label{fig_pretrain_maskratio}
   \end{minipage}
   \vspace{-1cm}
\end{figure}

In the main text, we have already mentioned that pre-training aids in the removal of various types of noise. We validated the relationship between pre-trained weights and different input noise intensities on the CSet dataset, as shown in Fig.~\ref{fig_pretrain_noiselevel}. Pre-training enhances denoising performance across different noise levels, particularly in the case of strong noise, where the knowledge provided by pre-training effectively avoids over-fitting to the noise.

\subsection{Masking Ratio}
We analyze the impact of pre-training on different paradigms under various masking ratios, as shown in Fig.~\ref{fig_pretrain_maskratio}. Our study reveals that pre-training plays a significant role in enhancing denoising performance across various masking ratios (especially in cases of $20 \leq p \leq 80$). 

\subsection{Discuss of Noise in Pre-training}
\label{sec:discuss_pretrain}
In the main text, we use the well-known natural image dataset ImageNet without making any assumptions about the presence or type of noise in each image, hoping to learn the statistical distribution rules from a large number of natural images. Here, we add synthetic noise of specified distribution and intensity during pre-training, and perform pre-training from noise to itself (denoted as ``+Gauss ($\sigma$=25) (N2N)''), and adopt the same iterative denoising strategy, proving that additional assumptions about specific noise type or noise level in pre-training leads to a decline in effectiveness, as shown in Table.~\ref{table_pretrain_noise}. Networks that are too small fail to learn sufficient denoising information, falling short of the corresponding zero-shot approaches.

\begin{table}[]
\vspace{-0.2cm}
\centering
\setlength{\abovecaptionskip}{0.1cm}
\setlength{\belowcaptionskip}{0.1cm}
\caption{Discussion of noise in pre-trained dataset. Additional assumptions of noise in pre-trained dataset result in lower performance ("+Gauss ($\sigma$=25)(N2N)").}
\label{table_pretrain_noise}
\setlength{\tabcolsep}{4mm}{
\resizebox{0.5\columnwidth}{!}{%
\begin{tabular}{lcc}
\toprule
Pre-train Mode                           & PSNR  & SSIM  \\ \midrule
\rowcolor[HTML]{EFEFEF} 
Ours                                     & 31.61 & 0.841 \\
+Gauss ($\sigma$=25)(N2N)      & 31.24 & 0.827 \\
\bottomrule
\end{tabular}%
}}
\vspace{-0.3cm}
\end{table}
\section{Extension to other network structures}
\label{sec:add_network}
In the main text, we discuss the effect of pre-training on the proposed model using the same network architecture as DIP~\cite{Ulyanov2018dip}. We are curious whether this prior knowledge could be applied to other model architectures. Here, we compare the impact of pre-training under different model settings. Specifically, we evaluate the removal of Gaussian noise with sigma=25 on CSet using three additional network architectures (DnCNN~\cite{zhang2017dncnn}, ResNet~\cite{he2016resnet}), as shown in Table~\ref{table_add_network}. The pre-training approach consistently brings performance gains across various network architectures.

\begin{table}[t]
\centering
\setlength{\abovecaptionskip}{0.1cm}
\setlength{\belowcaptionskip}{0.1cm}
\caption{Extension of proposed pre-training strategy into other network architectures. A performance improvement can be observed for both settings of beta=0.9 and 0.99 in the experiment across various network architectures.}
\label{table_add_network}
\setlength{\tabcolsep}{2mm}{
\resizebox{0.7\columnwidth}{!}{%
\begin{tabular}{cclccc}
\toprule
                         & Params (M)             & $\beta$    & Pre-trained   & Baseline    & Infer.time (s)                 \\ \midrule
                         &                        & 0.9  & \cellcolor[HTML]{EFEFEF}30.49/0.812 & 26.76/0.720 & 15.1           \\
\multirow{-2}{*}{DnCNN~\cite{zhang2017dncnn}}  & \multirow{-2}{*}{0.56} & 0.99 & \cellcolor[HTML]{EFEFEF}31.69/0.845 & 30.68/0.825 & 75.0          \\ \midrule
                         &                        & 0.9  & \cellcolor[HTML]{EFEFEF}30.46/0.812 & 29.20/0.778 &  8.0           \\
\multirow{-2}{*}{ResNet~\cite{he2016resnet}} & \multirow{-2}{*}{0.26} & 0.99 & \cellcolor[HTML]{EFEFEF}31.43/0.838 & 31.16/0.836 & 39.4           \\ \bottomrule
\end{tabular}%
}}
\vspace{-0.3cm}
\end{table}

\section{More Experimental Settings \& Results}
\label{sec:add_expset&result}
\subsection{Quantitative analysis of Poisson noise removal}
Due to space limitations in the text, there is no quantitative comparison of Poisson noise, which is listed in this section, see Table~\ref{table_poiss}.
\begin{table*}[tbp]
\centering
\setlength{\abovecaptionskip}{0.1cm}
\setlength{\belowcaptionskip}{0.1cm}
\caption{Quantitative comparison on CSet, McMaster and CBSD dataset for \textbf{Poisson noise removal} ($\lambda$$\in$[10,25,50]). For best results \textbf{highlighted} and second \underline{underlined}.}
\label{table_poiss}
\setlength{\tabcolsep}{1.5mm}{
\resizebox{1\columnwidth}{!}{%
\begin{tabular}{ccccccccc}
\toprule
 &
  $\lambda$ &
  DIP~\cite{Ulyanov2018dip} &
  N2V*~\cite{krull2019noise2void} &
  N2S*~\cite{batson2019noise2self} &
  ZS-N2N~\cite{Mansour2023zs-n2n} &
  FasterDIP~\cite{liu2023fasterdip} &
  Ours (faster) &
  Ours \\ \midrule
 &
  10 &
  22.88/0.495 &
  26.50/0.650 &
  25.34/0.661 &
  25.70/0.618 &
  25.02/0.633 &
  \cellcolor[HTML]{EFEFEF}\textbf{27.79/0.714} &
  \cellcolor[HTML]{EFEFEF}\underline{27.55/0.696} \\
 &
  25 &
  27.57/0.681 &
  27.16/0.755 &
  27.16/0.750 &
  28.06/0.711 &
  27.73/0.710 &
  \cellcolor[HTML]{EFEFEF}\underline{29.70/0.784} &
  \cellcolor[HTML]{EFEFEF}\textbf{30.02/0.785} \\
\multirow{-3}{*}{CSet~\cite{dabov2007bm3d}} &
  50 &
  30.03/0.775 &
  29.88/0.818 &
  27.68/0.780 &
  29.79/0.780 &
  28.86/0.749 &
  \cellcolor[HTML]{EFEFEF}\underline{31.00/0.830} &
  \cellcolor[HTML]{EFEFEF}\textbf{31.68/0.841} \\ \midrule
 &
  10 &
  24.45/0.644 &
  25.97/0.696 &
  25.68/0.735 &
  26.09/0.689 &
  26.14/0.730 &
  \cellcolor[HTML]{EFEFEF}\textbf{28.26/0.793} &
  \cellcolor[HTML]{EFEFEF}\underline{28.15/0.770} \\
 &
  25 &
  29.23/0.801 &
  28.84/0.807 &
  27.28/0.782 &
  28.49/0.775 &
  28.03/0.783 &
  \cellcolor[HTML]{EFEFEF}\underline{30.34/0.856} &
  \cellcolor[HTML]{EFEFEF}\textbf{30.92/0.862} \\
\multirow{-3}{*}{McMaster~\cite{ zhang2011mcmaster}} &
  50 &
  31.13/0.856 &
  30.60/0.871 &
  27.78/0.803 &
  30.34/0.834 &
  28.70/0.792 &
  \cellcolor[HTML]{EFEFEF}\underline{31.74/0.886} &
  \cellcolor[HTML]{EFEFEF}\textbf{32.64/0.900} \\ \midrule
 &
  10 &
  21.81/0.544 &
  23.17/0.686 &
  24.49/0.681 &
  25.25/0.662 &
  23.53/0.643 &
  \cellcolor[HTML]{EFEFEF}\textbf{26.27/0.716} &
  \cellcolor[HTML]{EFEFEF}\underline{26.05/0.709} \\
 &
  25 &
  26.83/0.741 &
  26.96/0.798 &
  26.20/0.775 &
  27.66/0.776 &
  26.64/0.757 &
  \cellcolor[HTML]{EFEFEF}\underline{28.91/0.823} &
  \cellcolor[HTML]{EFEFEF}\textbf{29.00/0.824} \\
\multirow{-3}{*}{CBSD~\cite{martin2001cbsd}} &
  50 &
  29.35/0.828 &
  28.25/0.835 &
  26.95/0.808 &
  29.55/0.833 &
  27.91/0.781 &
  \cellcolor[HTML]{EFEFEF}\underline{30.45/0.871} &
  \cellcolor[HTML]{EFEFEF}\textbf{31.00/0.881} \\ \bottomrule
\end{tabular}%
}}
\vspace{-0.4cm}
\end{table*}

\subsection{Quantitative comparison with more methods}
Here we compare more recent methods including additional DIP-based zero-shot method (DIP-SURE~\cite{jo2021rethinkingdip}), diffusion-based methods (DDNM~\cite{wang2022ddnm}, DDPG~\cite{garber2024ddpg}), and zero-shot modifications from unsupervised methods (AP-BSN~\cite{lee2022ap-bsn}, MM-BSN~\cite{zhang2023mm-bsn}, PUCA~\cite{jang2024puca}), refer to Table~\ref{rebuttal_tab_comp} for results.

Specifically, for DIP-SURE, we report both peak performance and the performance at the final iteration. Since DIP-SURE is specifically designed for Gaussian and Poisson noise, and requires the input of Gaussian noise variance, for real-world denoising tasks we provide estimated variance from paired data, to report the best performance. For diffusion-based methods, which are trained exclusively on Gaussian noise and also require variance as prior, we use the same variance estimation approach to report their best results. For self-supervised methods, which can be easily adapted to a single image with minimal changes, we follow their original settings. In each iteration, we crop eight same-size patches from the noisy image to form a batch, and perform inference on the full image every 10 iterations, and combine denoised images using the same ensemble strategy as our method for fairness.

We observe that DIP-SURE, due to its priors on noise type and variance, performs slightly better than our method under Gaussian noise settings. However, its performance significantly drops when dealing with real noise, especially when reporting the last performance. Since diffusion models are inherently Gaussian denoisers, they perform well on Gaussian noise when the variance is known, but also face challenges with real noise. The modified blind-spot network-based methods can handle severe real noise relatively well, but they may suffer from potential image detail loss and require long inference times.

\subsection{Ensemble results of N2V and N2S}
In the main text, we present versions of DIP~\cite{Ulyanov2018dip} and FasterDIP~\cite{liu2023fasterdip} with EMA (Exponential Moving Average) ensembles, as these processes are included in their source codes. To provide additional information for comparison, we also adapted N2V*~\cite{krull2019noise2void} and N2S*~\cite{batson2019noise2self} to their corresponding EMA ensemble versions, as shown in Table~\ref{table_n2s_n2v_ema_gauss},~\ref{table_n2s_n2v_ema_general},~\ref{table_n2s_n2v_ema_real}. Generally, the ensemble versions of these methods can improve the PSNR by 1$\sim$2 dB. However, even though the enhanced N2V may outperform our Faster version in some cases, it does not affect the performance comparison with our $\beta$=0.99 version, which remains the best. Moreover, our $\beta$=0.99 version achieves this with less than half the inference time required by these methods.

\begin{table*}[]
\setlength{\abovecaptionskip}{0.1cm}
\setlength{\belowcaptionskip}{0.1cm}
\caption{Quantitative comparison with other DIP-based method (DIP-SURE~\cite{jo2021rethinkingdip}), Diffusion methods (DDNM, DDPG), zero-shot methods modified from other Self-supervised methods (AP-BSN, MM-BSN, PUCA). DIP-SURE, DDNM and DDPG requires additional noise variance as input, and DIP-SURE applies different iterations for each image.}
\setlength{\tabcolsep}{1.5mm}{
\resizebox{1\columnwidth}{!}{%
\begin{tabular}{@{}llccccccc@{}}
\toprule
\multicolumn{2}{c}{\multirow{2}{*}{Method}} & \multicolumn{3}{c}{CSet+Gaussian} & \multirow{2}{*}{\begin{tabular}[c]{@{}c@{}}SIDD\\ validation\end{tabular}} & \multirow{2}{*}{PolyU} & \multirow{2}{*}{FMD} & \multirow{2}{*}{\begin{tabular}[c]{@{}c@{}}Avg. Infer.\\ time (s)\end{tabular}} \\ \cmidrule(lr){3-5}
\multicolumn{2}{c}{} & $\sigma$=10 & $\sigma$=25 & $\sigma$=50 &  &  &  &  \\ \midrule
\multirow{2}{*}{DIP} & DIP-SURE(peak) & 35.37/0.916 & 31.88/0.855 & 28.81/0.775 & 30.45/0.727 & 35.87/0.944 & 32.04/0.798 & - \\
 & DIP-SURE(last) & 34.98/0.908 & 31.50/0.840 & 28.76/0.762 & 26.63/0.649 & 35.78/0.942 & 32.07/0.793 & 367.3 \\ \midrule
\multirow{2}{*}{Diff.} & DDNM & 36.22/0.927 & 32.4/0.859 & 29.99/0.793 & 28.11/0.597 & 37.15/0.935 & 28.99/0.685 & 26.7 \\
 & DDPG & 32.43/0.826 & 27.07/0.606 & 15.95/0.183 & 29.84/0.612 & 35.79/0.887 & 30.41/0.735 & 24.3 \\ \midrule
\multirow{3}{*}{\begin{tabular}[c]{@{}l@{}}Self-\\ supervised\end{tabular}} & AP-BSN* & - & 25.04/0.671 & - & \multicolumn{1}{l}{33.34/0.847} & 32.64/0.928 & 29.27/0.799 & 351.4 \\
 & MM-BSN* & - & 25.27/0.676 & - & 33.36/0.843 & 33.07/0.930 & 29.73/0.810 & 505.3 \\
 & PUCA* & - & 24.74/0.640 & - & 33.52/0.816 & 33.31/0.927 & 30.22/0.808 & 450.0 \\ \midrule
 & Ours & 34.91/0.909 & 31.61/0.841 & 28.26/0.710 & 34.43/0.844 & 38.11/0.962 & 32.97/0.847 & 45.8 \\ \bottomrule
 \label{rebuttal_tab_comp}
\end{tabular}
}}
\end{table*}
\begin{figure*}[tbp]
\setlength{\abovecaptionskip}{0.1cm}
\setlength{\belowcaptionskip}{0.1cm}
  \centering
   \includegraphics[width=1\linewidth]{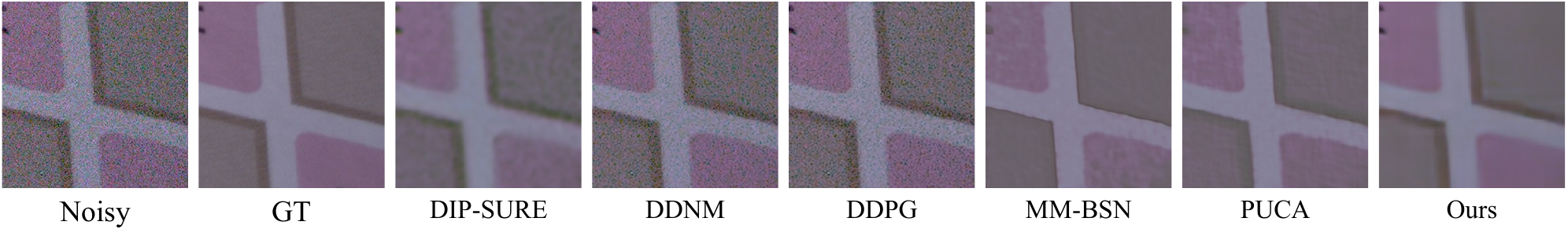}
   \caption{Comparison of different methods under SIDD validation (SIDDval\_34\_22).}
   \vspace{-0.2cm}
   \label{visual_sidd_rebuttal}
\end{figure*}
\begin{table}[t]
\centering
\setlength{\abovecaptionskip}{0.1cm}
\setlength{\belowcaptionskip}{0.1cm}
\caption{Quantitative comparison of ensemble version of N2V* and N2S* on CSet, McMaster and CBSD dataset for Gaussian and Poisson noise removal.}
\label{table_n2s_n2v_ema_gauss}
\setlength{\tabcolsep}{1.7mm}{
\resizebox{\columnwidth}{!}{%
\begin{tabular}{clccc|ccc}
\toprule
 &  & \multicolumn{3}{c|}{Gaussian} & \multicolumn{3}{c}{Poisson} \\ \cmidrule(lr){3-8}
                      &          & $\beta$=10  & $\beta$=25  & $\beta$=50  & $\lambda$=10 & $\lambda$=25 & $\lambda$=50 \\ \midrule
\multirow{3}{*}{N2V*~\cite{krull2019noise2void}} & CSet~\cite{dabov2007bm3d}     & 34.05/0.895 & 30.99/0.825 & 27.70/0.703 & 27.02/0.675  & 29.55/0.790  & 30.95/0.830   \\
                      & McMaster~\cite{ zhang2011mcmaster} & 34.31/0.920 & 30.97/0.862 & 27.94/0.766 & 26.93/0.751  & 30.06/0.812  & 31.78/0.884                        \\
                      & CBSD~\cite{martin2001cbsd}     & 33.15/0.918 & 30.05/0.850 & 26.27/0.702 & 25.42/0.714  & 28.55/0.811  & 30.40/0.869                        \\ \midrule
\multirow{3}{*}{N2S*~\cite{batson2019noise2self}
} & CSet~\cite{dabov2007bm3d}     & 29.92/0.843 & 28.76/0.787 & 26.67/0.704 & 26.83/0.670  & 27.94/0.765  & 28.91/0.797 \\
                      & McMaster~\cite{ zhang2011mcmaster} & 29.85/0.867 & 28.42/0.785 & 25.07/0.678 & 26.35/0.760  & 28.24/0.814  & 28.99/0.836                        \\
                      & CBSD~\cite{martin2001cbsd}     & 28.50/0.854 & 27.51/0.803 & 25.07/0.705 & 24.83/0.714  & 26.59/0.785  & 27.53/0.817                        \\ \bottomrule
\end{tabular}%
}}
\end{table}
\begin{table}[t]
\centering
\setlength{\abovecaptionskip}{0.1cm}
\setlength{\belowcaptionskip}{0.1cm}
\caption{Quantitative comparison of ensemble version of N2V* and N2S* on Kodak with 5 noise types for generalization evaluation.}
\label{table_n2s_n2v_ema_general}
\setlength{\tabcolsep}{1.3mm}{
\resizebox{\columnwidth}{!}{%
\begin{tabular}{ccccccc|c}
\toprule
\multirow{2}{*}{} &
  \multicolumn{2}{c}{Gaussian} &
  \multirow{2}{*}{\begin{tabular}[c]{@{}c@{}}Poisson\\ $\lambda$$\in$[10,50]\end{tabular}} &
  \multirow{2}{*}{NLF} &
  \multirow{2}{*}{\begin{tabular}[c]{@{}c@{}}Speckle\\ $v$$\in$[10,50]\end{tabular}} &
  \multirow{2}{*}{\begin{tabular}[c]{@{}c@{}}S\&P\\ $d$$\in$[0.02,0.05]\end{tabular}} &
  \multirow{2}{*}{Average} \\
   \cline{2-3}  & \raisebox{-0.3ex}{$\sigma$=25} & \raisebox{-0.3ex}{$\sigma$$\in$[10,50]} &             &             &             &             &             \\ \midrule
N2V* & 30.95/0.850 & 30.90/0.836          & 29.55/0.825 & 32.26/0.888 & 33.82/0.917 & 34.26/0.948 & 31.95/0.877 \\
N2S* & 28.34/0.804 & 27.91/0.790          & 27.21/0.779 & 28.82/0.840 & 28.84/0.846 & 28.88/0.828 & 28.33/0.816 \\ \bottomrule
\end{tabular}%
}}
\end{table}
\begin{table}[t]
\centering
\setlength{\abovecaptionskip}{0.1cm}
\setlength{\belowcaptionskip}{0.1cm}
\caption{Quantitative comparison of ensemble version of N2V* and N2S* on SIDD, PolyU and FMD dataset for real-world noise removal.}
\label{table_n2s_n2v_ema_real}
\setlength{\tabcolsep}{4mm}{
\resizebox{0.8\columnwidth}{!}{%
\begin{tabular}{lcccc}
\toprule
\multirow{2}{*}{Methods} & \multicolumn{2}{c}{SIDD~\cite{abdelhamed2018sidd}} & \multirow{2}{*}{PolyU~\cite{xu2018polyu}} & \multirow{2}{*}{FMD~\cite{zhang2019fmd}} \\
\cline{2-3} & \raisebox{-0.3ex}{validation} & \raisebox{-0.5ex}{benchmark} &  &  \\
\midrule
N2V*~\cite{krull2019noise2void} & 28.51/0.670 & 27.19/0.645 & 36.11/0.921 & 30.85/0.754 \\
N2S*~\cite{batson2019noise2self}
 & 27.41/0.584 & 27.59/0.684 & 35.39/0.939 & 31.72/0.762 \\ \bottomrule
\end{tabular}%
}}
\end{table}

\subsection{Details of Unseen Noise}
\label{sec:noise_general}
\noindent\textbf{Gaussian Noise.}
Gaussian noise follows a normal distribution and is commonly encountered in digital imaging, especially during sensor data acquisition and transmission. It represents random variations in intensity and color information in images, making it a fundamental noise model in image processing. For each element in clean image $I_{[k]}$, is represented by:
\begin{equation}
\hat{I}_{[k]} = I_{[k]} + \sigma \cdot N_{[k]},
\end{equation}
where $N_{[k]}$ represents random variable sampled from a standard normal distribution, which is characterized by its standard deviation (\(\sigma\)).

\noindent\textbf{Poisson Noise.}
Poisson noise is prevalent in scenarios with low-light conditions, such as astronomical imaging or medical imaging, where the photon count is inherently random and follows a Poisson distribution. Poisson noise models the variation of intensity based on a Poisson distribution, is generally expressed as:
\begin{equation}
\hat{I}_{[k]} = P(I_{[k]} \cdot \lambda)/\lambda,
\end{equation}
where \( \lambda \) indicates the event rate, and $P(\cdot)$ denotes random variable generated from Poisson distribution.

\noindent\textbf{Noise Level Function (NLF).}
Noise level function, also referred to as heteroscedastic gaussian model~\cite{abdelhamed2019noiseflow}, is commonly described by a varying standard deviation across the image. This type of noise is widely used to express the read-shot noise in camera imaging pipeline, where different parts of the image exhibit different noise levels. It is typically modeled as:
\begin{equation}
\hat{I}_{[k]}\sim\mathcal{N}(\mu={I}_{[k]}, \sigma^2=\sigma_r + \sigma_s \cdot I_{[k]})
\end{equation}
where \(\sigma_r\) and \(\sigma_r\) represent different standard deviations in distinct regions of the image. Noise parameter calibrated for~\cite{plotz2017darmstadt} in work~\cite{brooks2019unprocessing} obeys a log-linear rule:
\begin{equation}
log(\sigma_r)=2.18\cdot log(\sigma_s)+1.2
\end{equation}
We choose $\sigma_s\in [0.01,0.012]$ to better illustrate the generality.

\noindent\textbf{Speckle Noise.}
Speckle noise is an interference pattern produced by the coherent processing of a signal, especially in active radar and ultrasound imaging. This noise is particularly common in radar and ultrasound images, where it can significantly degrade the quality of the image. Its mathematical representation is:
\begin{equation}
\hat{I}_{[k]} = I_{[k]} + I_{[k]} \cdot U_{[k]},
\end{equation}
where $U_{[k]}$ is sampled from uniform distribution with mean 0 and \(v\) representing the standard deviation of the noise.

\noindent\textbf{Salt-and-Pepper Noise (S\&P).}
Salt-and-Pepper noise, also known as impulse noise, is characterized by sharp and sudden disturbances in an image signal. It's typically represented as sparse white and black pixels, hence the name. This noise can be caused by sharp and sudden disturbances in the image signal, often due to transmission errors, faulty memory locations, or timing errors in digital image sensors. Its mathematical representation is:
\begin{equation}
\hat{I}_{[k]} = I_{[k]} + S_{[k]} - S_{[k]},
\end{equation}
where $S_{[k]}$ and $P_{[k]}$ represents salt and pepper noise, respectively. For each of them are Bernoulli sample with probability $d$ of $I_{max}/I_{min}$ and probability $1-d$ of 0, which makes the probability total affected is $2\cdot d$.

\subsection{Additional Computational Costs}
\label{sec:comp_costs}
In analyzing the performance of deep learning models, it's crucial to consider both the Floating Point Operations (FLOPs) and the model parameters. FLOPs give us an insight into the computational complexity of the model, which affects inference time and resource utilization, while the number of parameters indicates its capacity to learn and adapt to complex data patterns. A balance between these two aspects is essential for efficient and effective model performance. Our analysis, as reflected in the comparison between Table~\ref{table_flops}, demonstrates that our method successfully achieves this balance. It maintains computational efficiency without compromising the model's ability to accurately process and analyze data, an essential factor for practical application in varied computational environments.

\begin{table}[]
\centering
\setlength{\abovecaptionskip}{0.1cm}
\setlength{\belowcaptionskip}{0.1cm}
\caption{Efficiency comparisons of deep learning-based
methods on Params and FLOPs under input size $256\times 256$ with a single forward step. Iterations used for synthetic noise is provided for reference.}
\label{table_flops}
\setlength{\tabcolsep}{4mm}{
\resizebox{0.6\columnwidth}{!}{%
\begin{tabular}{c|ccc}
\toprule
Method       & Params (M)     & FLOPs (G)    & Iters  \\ \midrule
DIP~\cite{Ulyanov2018dip}          & 2.3            & 19.66        & 3,000 \\
N2V*~\cite{krull2019noise2void}          & 1.2            & 80.50        & 1,500 \\
N2S*~\cite{batson2019noise2self}          & 0.07           & 1.57         & 1,800 \\
ZS-N2N~\cite{Mansour2023zs-n2n}       & 0.02           & 1.45         & 2,000 \\
FasterDIP~\cite{liu2023fasterdip}    & 0.05$\sim$0.92 & 0.5$\sim$8.8 & 3,000 \\
\rowcolor[HTML]{EFEFEF} 
Ours(faster) & 0.73           & 8.11         & 1,000 \\
\rowcolor[HTML]{EFEFEF} 
Ours         & 0.73           & 8.11         & 200 \\ \bottomrule
\end{tabular}%
}}
\end{table}

\subsection{Zero-shot Denoising on More Image Types}
\begin{figure}[h]
\setlength{\abovecaptionskip}{0.0cm}
\setlength{\belowcaptionskip}{0.0cm}
  \centering
   \includegraphics[width=0.8\linewidth]{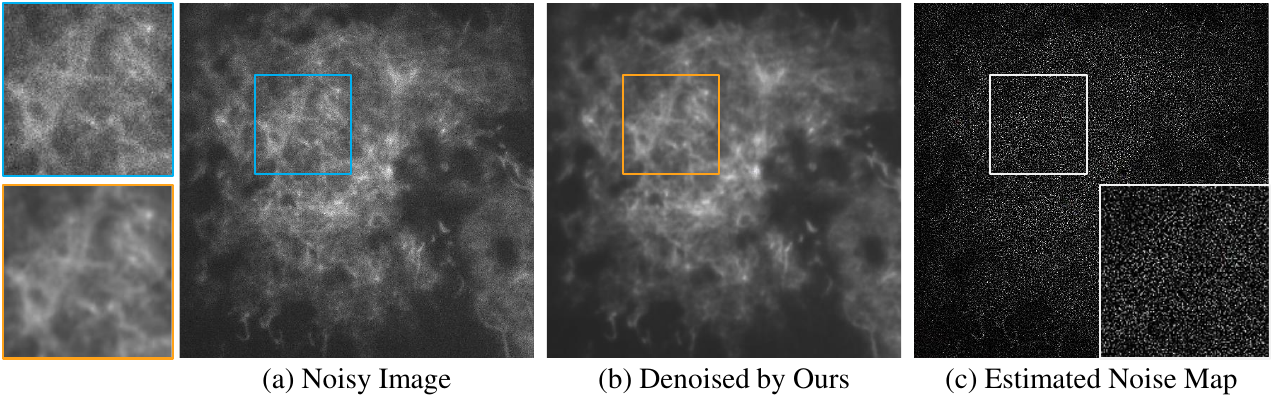}
  \caption{On a noisy microscopy image (a) using the proposed MPI to denoise retaining the structural information in the image as much as possible (b), see the noise map (c).}
   \label{fig_micro}
\end{figure}

\begin{figure}[h]
\setlength{\abovecaptionskip}{0.0cm}
\setlength{\belowcaptionskip}{0.0cm}
  \centering
   \includegraphics[width=0.8\linewidth]{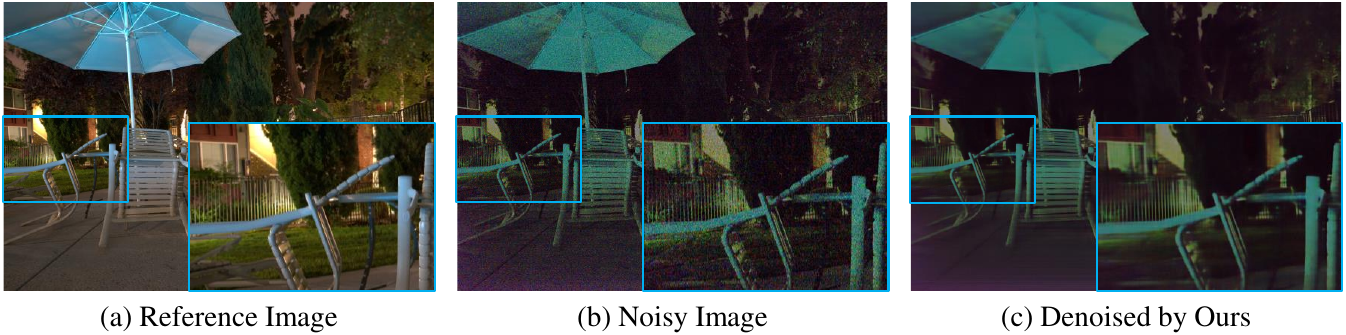}
  \caption{For extremely low-light images, there is serious color bias in the expected denoising result (a) and captured noisy image (b). This color bias is still retained in the denoising result, but the noise is basically removed. This is discussed in Sec.~\ref{sec_limitations}.}
   \label{fig_lowlight}
\end{figure}

In the main text, we demonstrate the ability of proposed MPI to generalize to other types of images through a medical imaging dataset. Further here, we explore new types of images, including a microscopy imaging dataset BioSR~\cite{qiao2021biosr} and extremely low-light dataset SID~\cite{chen2018sid}. See Fig.~\ref{fig_micro} and Fig.~\ref{fig_lowlight} for qualitative examples.

\section{Concluding Remarks}
In this study, we introduce Masked Pre-train then Iterative fill (MPI), a zero-shot denoising paradigm utilizing pre-trained model with random masks on natural images.
The pre-trained weights is optimized on a specific noisy image through Iterative filling process, and predictions with corresponding masks during inference is combined for enhanced quality and faster inference.

\subsection{Broader impacts}
From the perspective of our work, we have pioneered the use of generalizable knowledge from natural images without any assumptions about noise degradation, offering an efficient framework for handling diverse synthetic and real noises with significantly reduced inference time, which is a critical issue in zero-shot denoising and makes their practical applications feasible. Notably, our zero-shot method excels in generalization compared to current supervised and unsupervised methods, offering new insights into denoising.

\subsection{Limitations}
\label{sec_limitations}
Although proposed MPI has shown effects in removal of various types of noises, the mask-based noise-supervised denoising setting does not seem to allow the removal of non-zero mean noise. So when dealing with extremely low-light images with severe color bias, the color bias still remains after denoising; this is a common problem in zero-shot denoising, because there is no prior regarding noise-clean image pair in specific domain, but it may limit several practical applications, and we are currently trying to solve this problem in other ways.
\newpage
\section{Additional Qualitative Results}
\label{sec:add_qualitative}
\begin{figure*}[h]
\begin{center}
\includegraphics[width=1\textwidth]{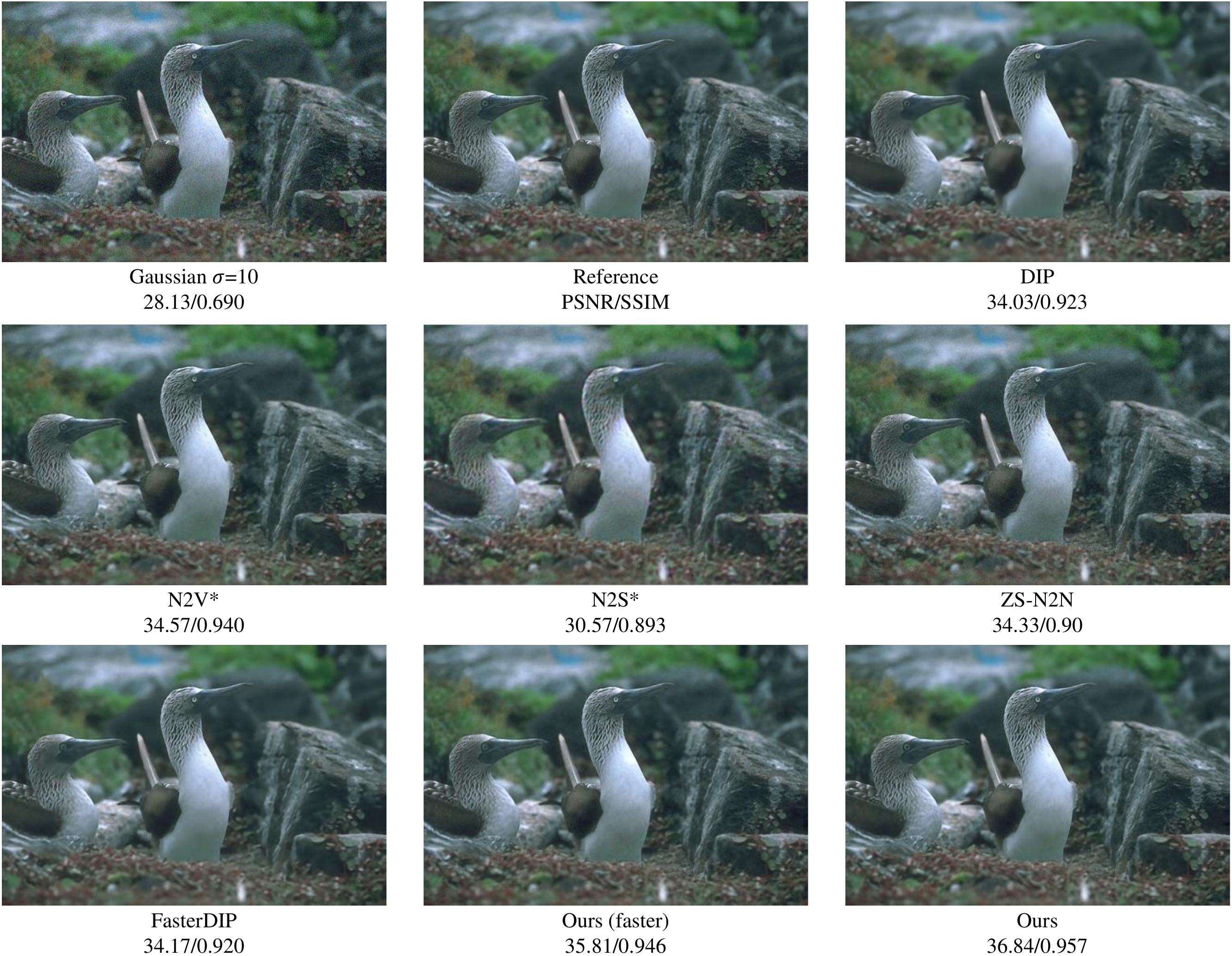}
\end{center}
\caption{Qualitative comparison of results on CBSD~\cite{martin2001cbsd} with Gaussian $\sigma$=10. Noisy patch is from CBSD-11.
}
\label{fig_gauss_10}
\end{figure*}

\begin{figure*}
\begin{center}
\includegraphics[width=1\textwidth]{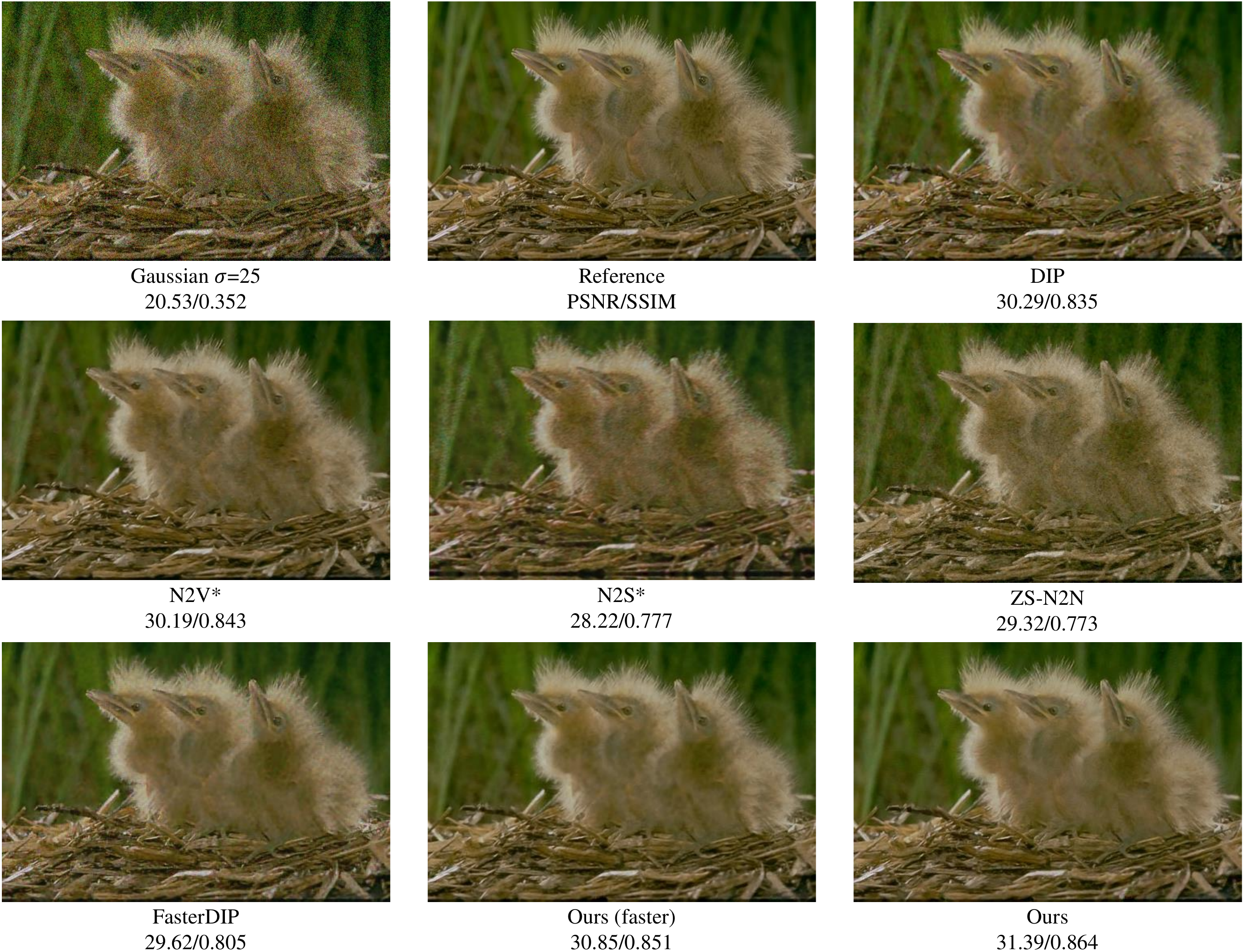}
\end{center}
\caption{Qualitative comparison of results on CBSD~\cite{martin2001cbsd} with Gaussian $\sigma$=25. Noisy patch is from CBSD-31.
}
\label{fig_gauss_25}
\end{figure*}

\begin{figure*}
\begin{center}
\includegraphics[width=1\textwidth]{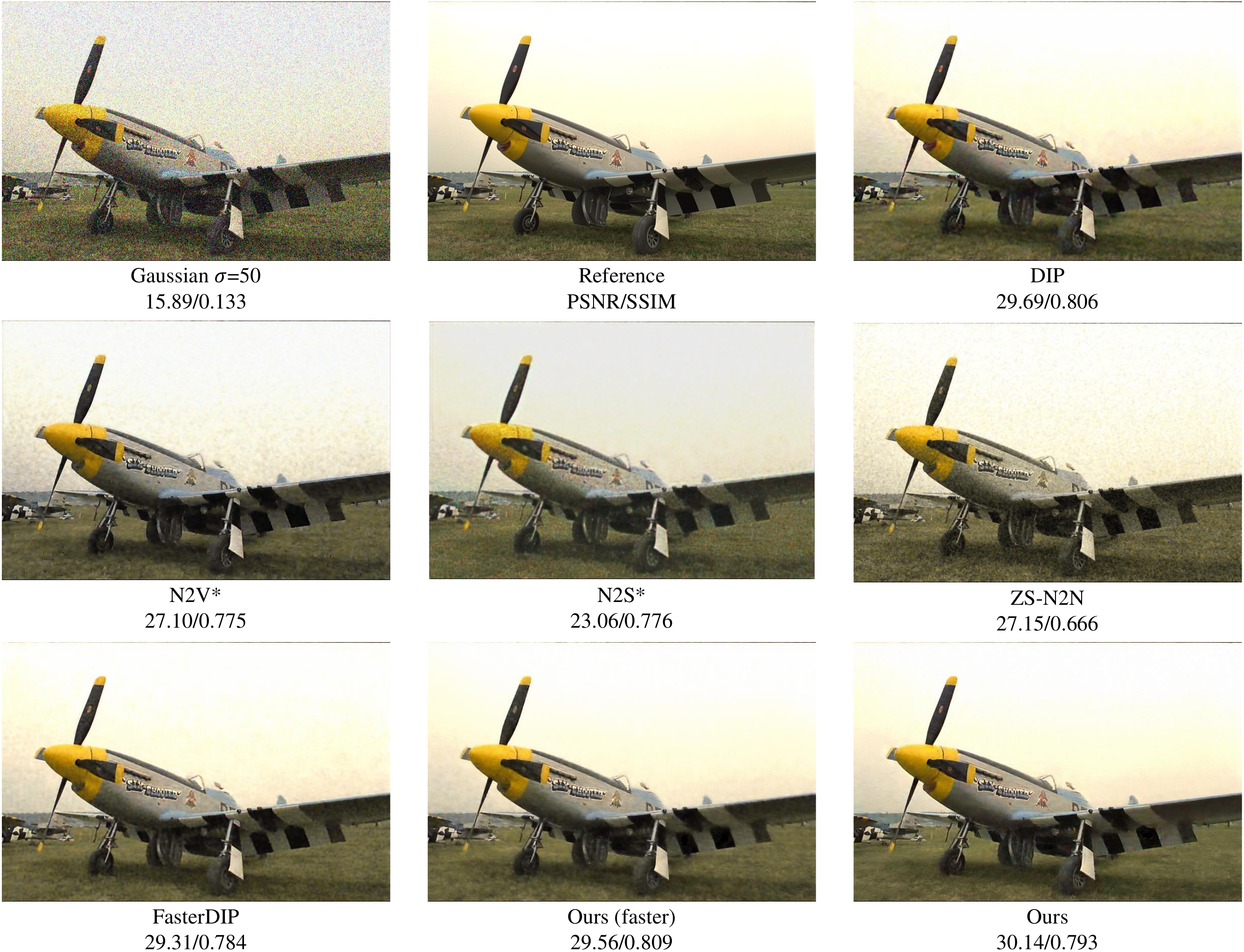}
\end{center}
\caption{Qualitative comparison of results on Kodak~\cite{franzen1999kodak} with Gaussian $\sigma$=50. Noisy patch is from kodim20.
}
\label{fig_gauss_50}
\end{figure*}

\begin{figure*}
\begin{center}
\includegraphics[width=0.8\textwidth]{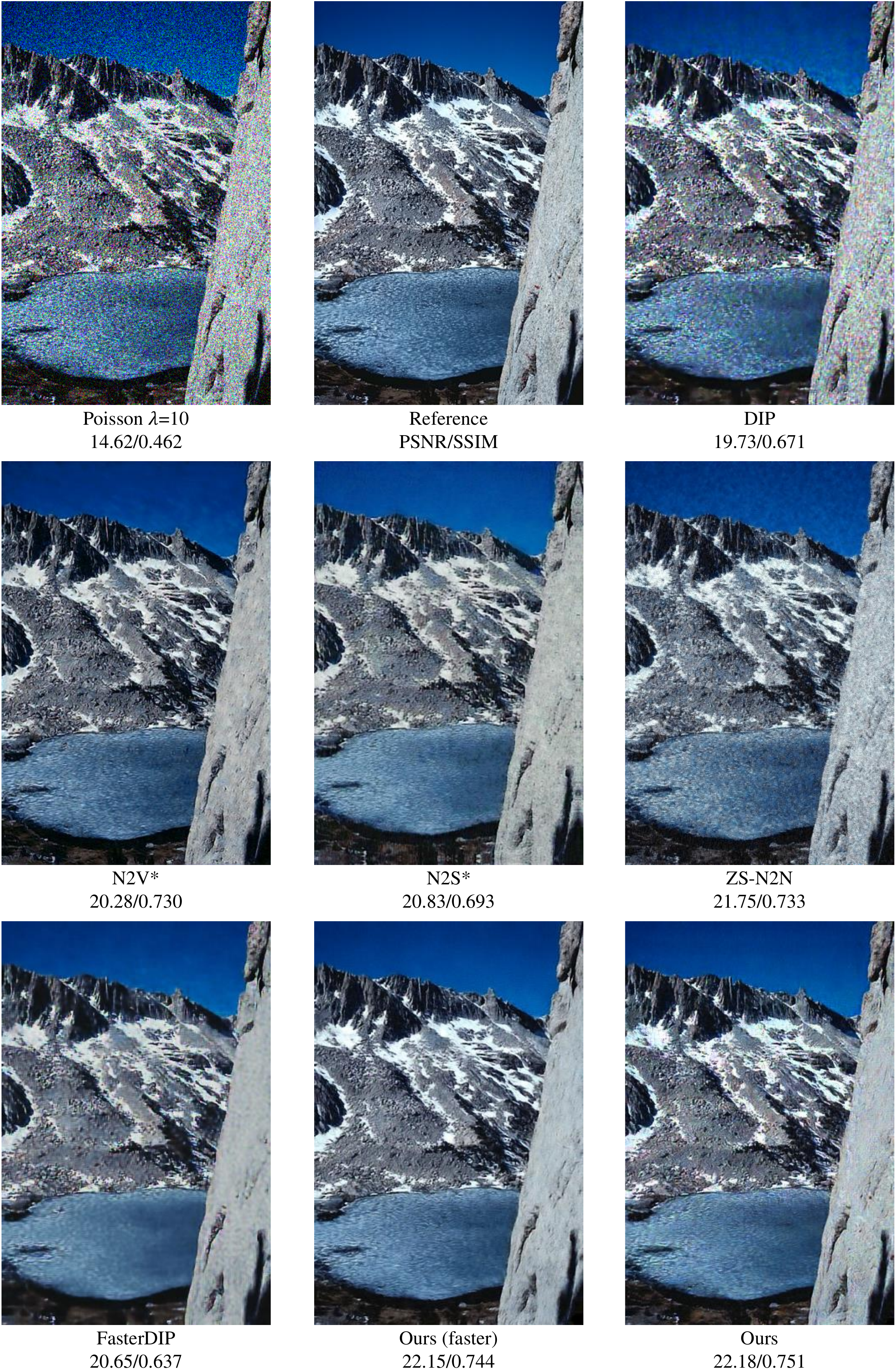}
\end{center}
\caption{Qualitative comparison of results on CBSD~\cite{martin2001cbsd} with Poisson $\lambda$=10. Noisy patch is from CBSD-33.
}
\label{fig_poiss_10}
\end{figure*}

\begin{figure*}
\begin{center}
\includegraphics[width=0.8\textwidth]{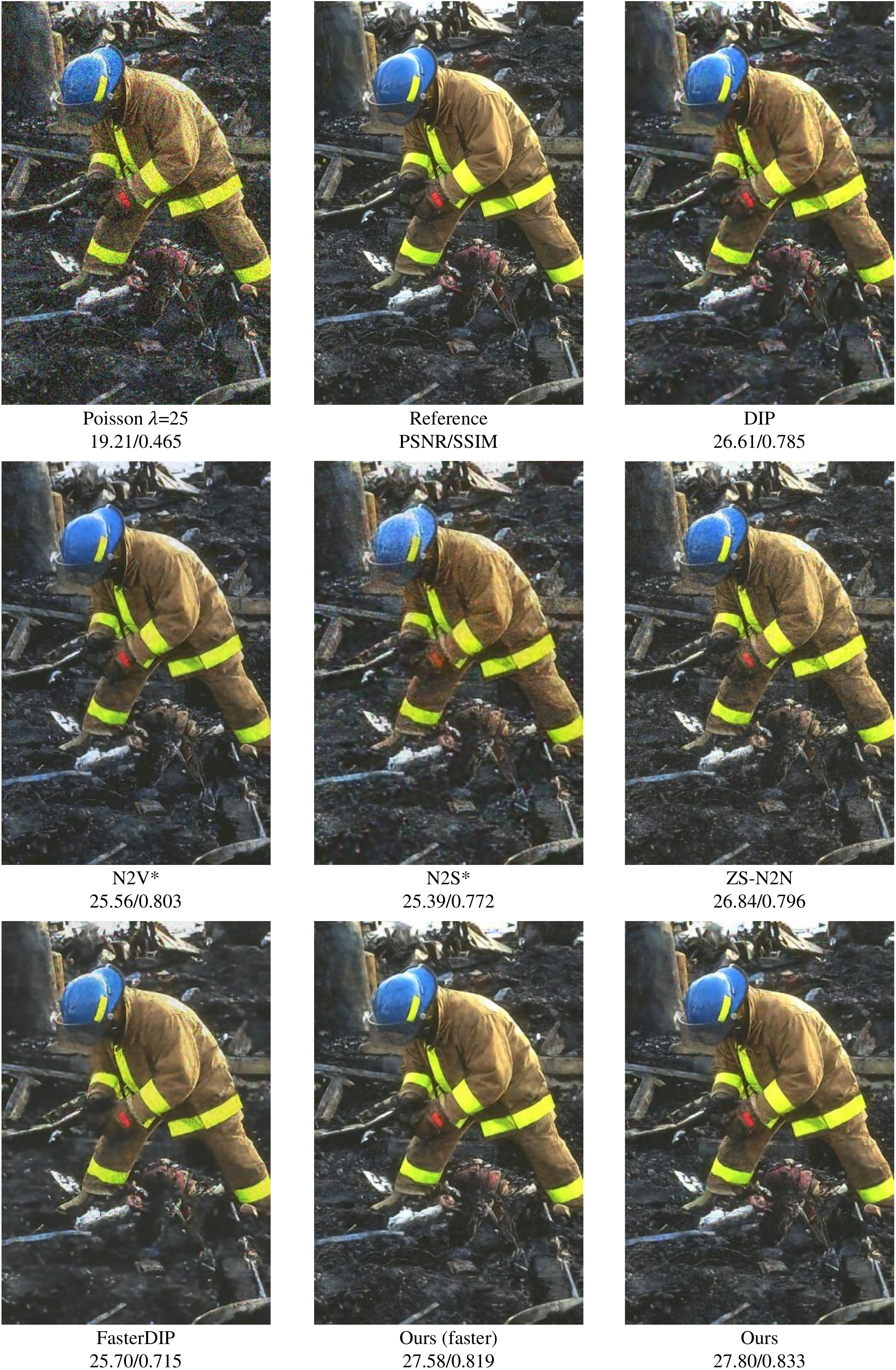}
\end{center}
\caption{Qualitative comparison of results on CBSD~\cite{martin2001cbsd} with Poisson $\lambda$=25. Noisy patch is from CBSD-56.
}
\label{fig_poiss_25}
\end{figure*}

\begin{figure*}
\begin{center}
\includegraphics[width=1\textwidth]{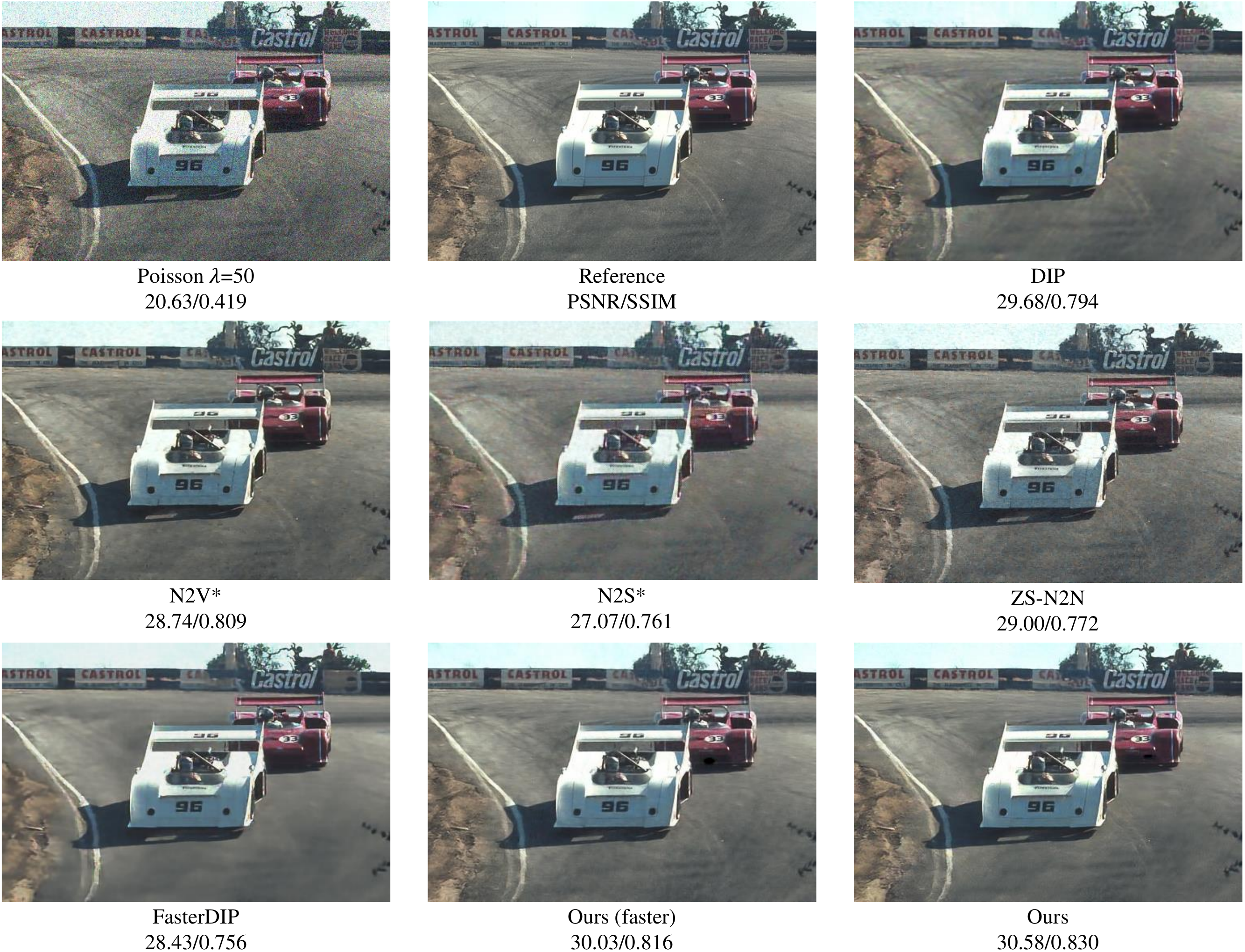}
\end{center}
\caption{Qualitative comparison of results on CBSD~\cite{martin2001cbsd} with Poisson $\lambda$=50. Noisy patch is from CBSD-05.
}
\label{fig_poiss_50}
\end{figure*}

\begin{figure*}
\begin{center}
\includegraphics[width=1\textwidth]{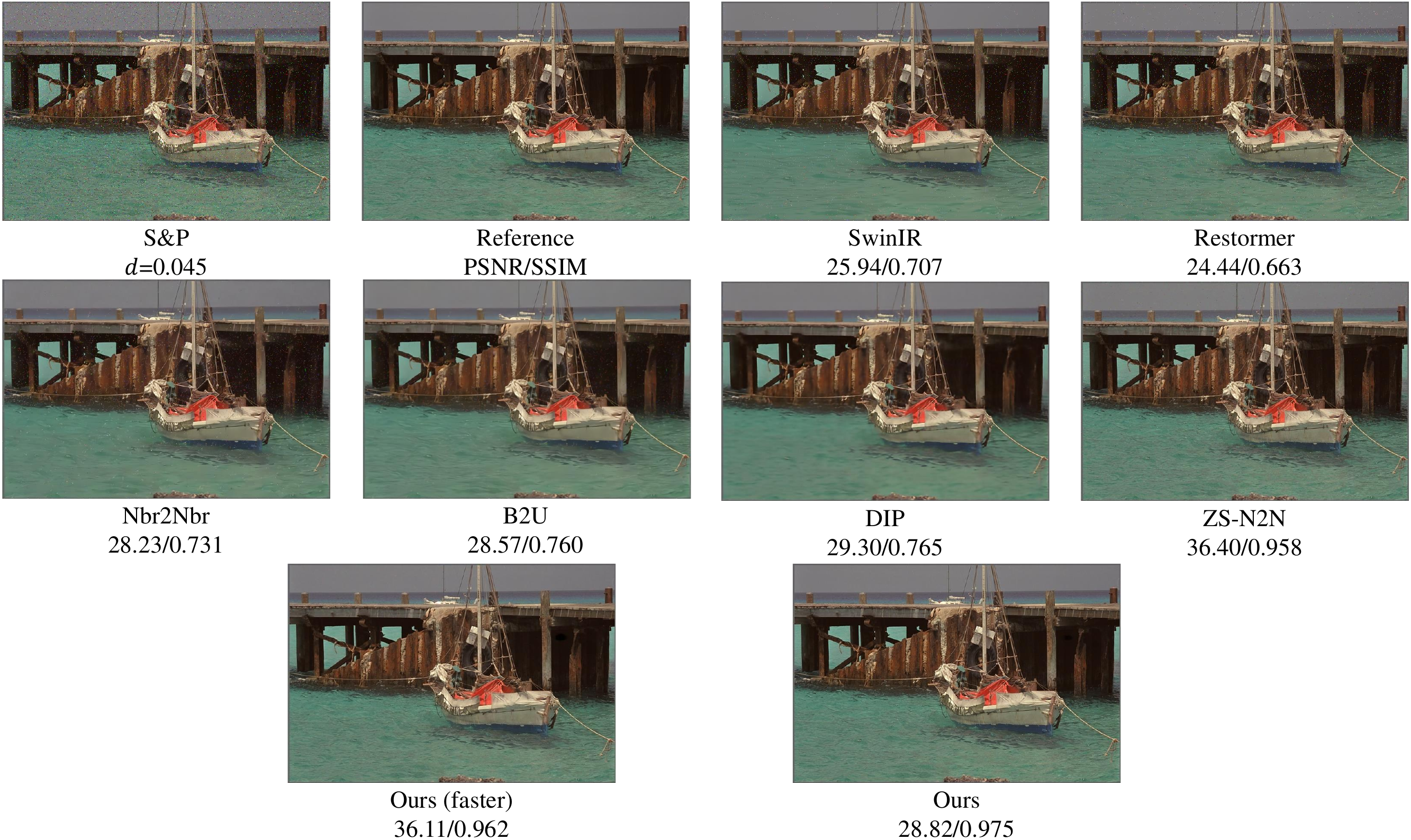}
\end{center}
\caption{Qualitative comparison of generalization on Kodak~\cite{franzen1999kodak} with S\&P $d$=0.045. Noisy patch is from kodim11.
}
\label{fig_general_s&p}
\end{figure*}

\begin{figure*}
\begin{center}
\includegraphics[width=1\textwidth]{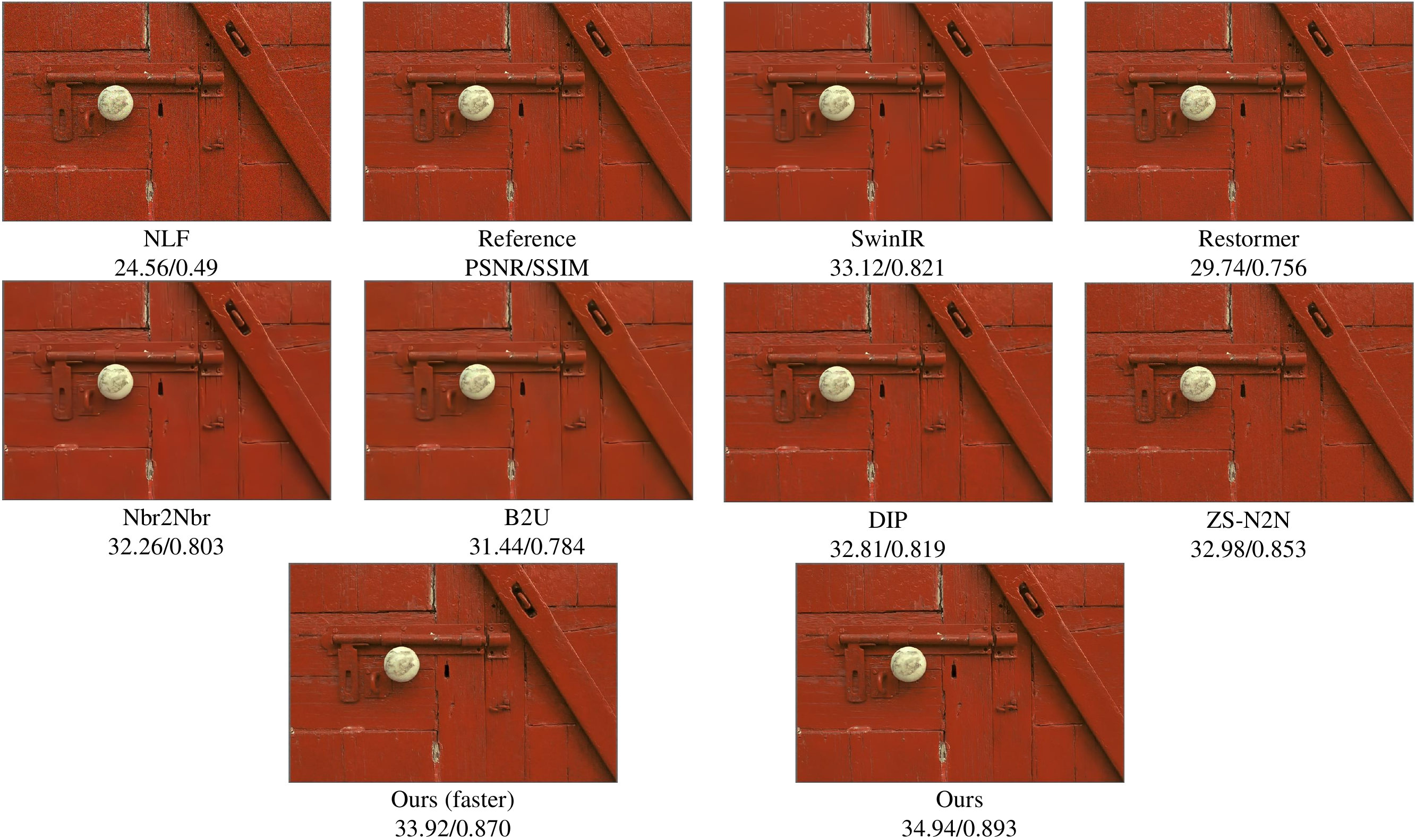}
\end{center}
\caption{Qualitative comparison of generalization on Kodak~\cite{franzen1999kodak} with NLF~\cite{plotz2017darmstadt}. Noisy patch is from kodim02.
}
\label{fig_general_nlf}
\end{figure*}

\begin{figure*}
\begin{center}
\includegraphics[width=1\textwidth]{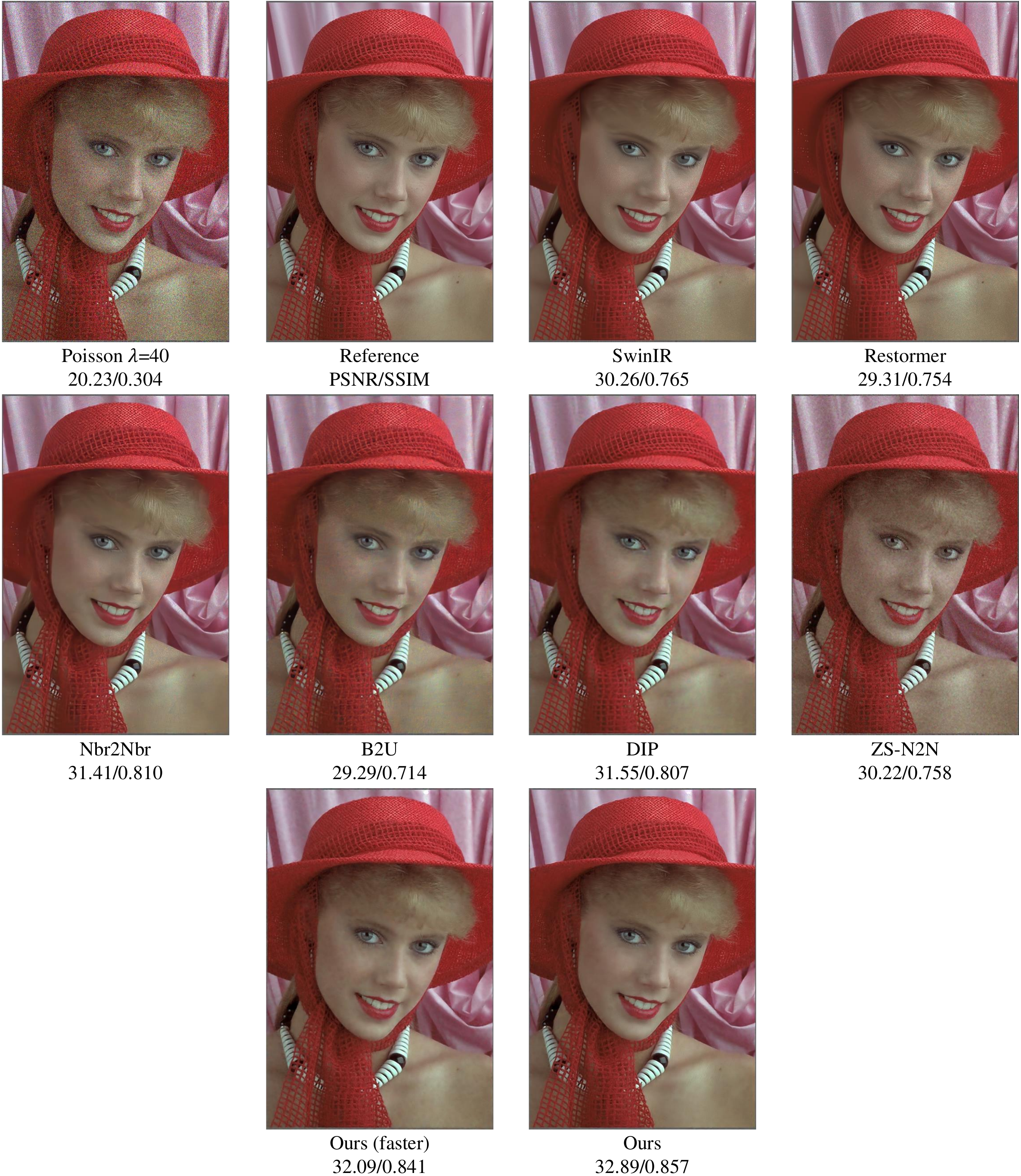}
\end{center}
\caption{Qualitative comparison of generalization on Kodak~\cite{franzen1999kodak} with Poisson $\lambda$=40. Noisy patch is from kodim04.
}
\label{fig_general_poiss}
\end{figure*}

\begin{figure*}
\begin{center}
\includegraphics[width=0.8\textwidth]{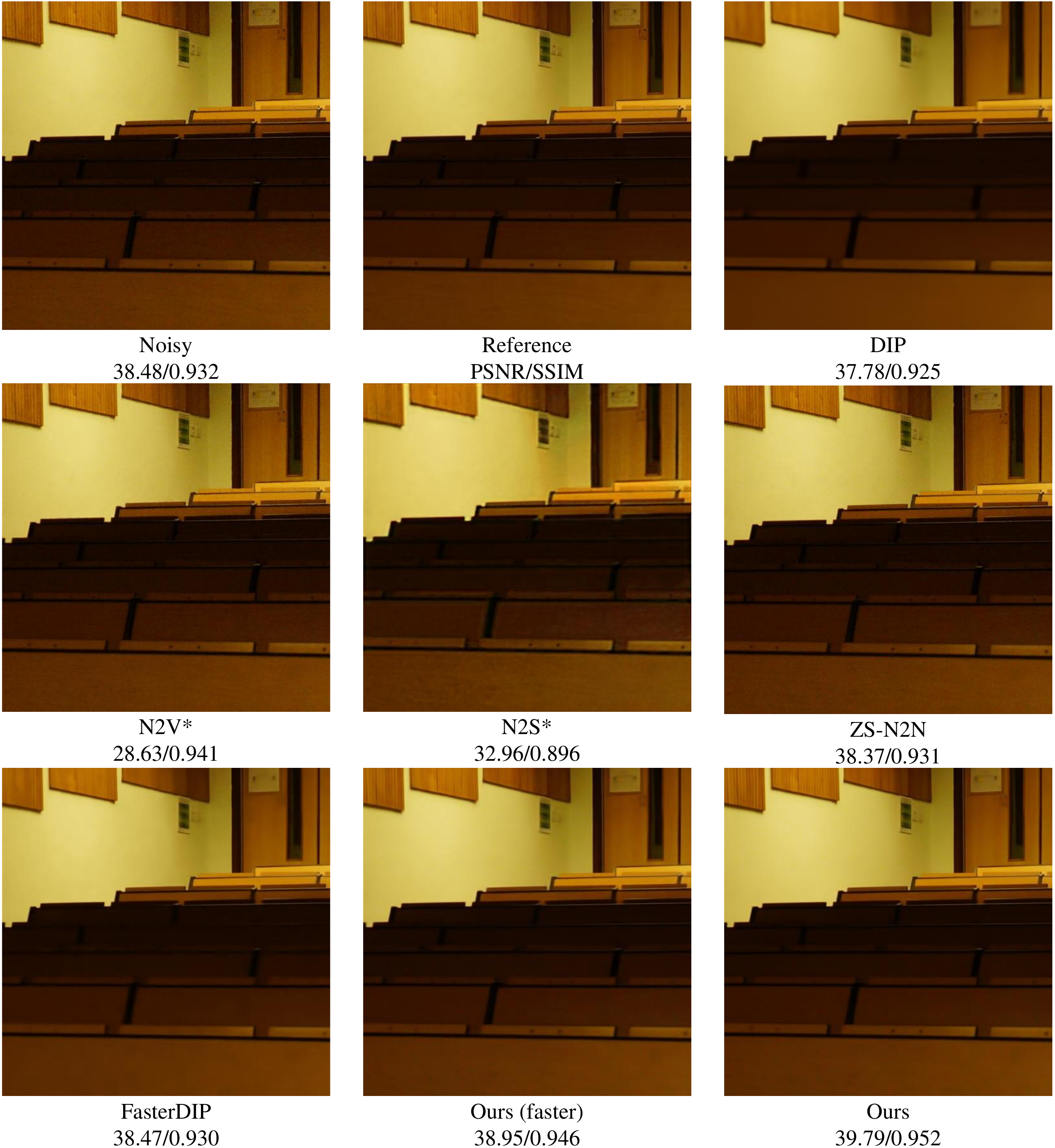}
\end{center}
\caption{Qualitative comparison of realnoise on PolyU~\cite{xu2018polyu}. Noisy patch is from Sony\_3-5\_200\_1600\_classroom\_14.
}
\label{fig_general_polyu1}
\end{figure*}

\begin{figure*}
\begin{center}
\includegraphics[width=0.8\textwidth]{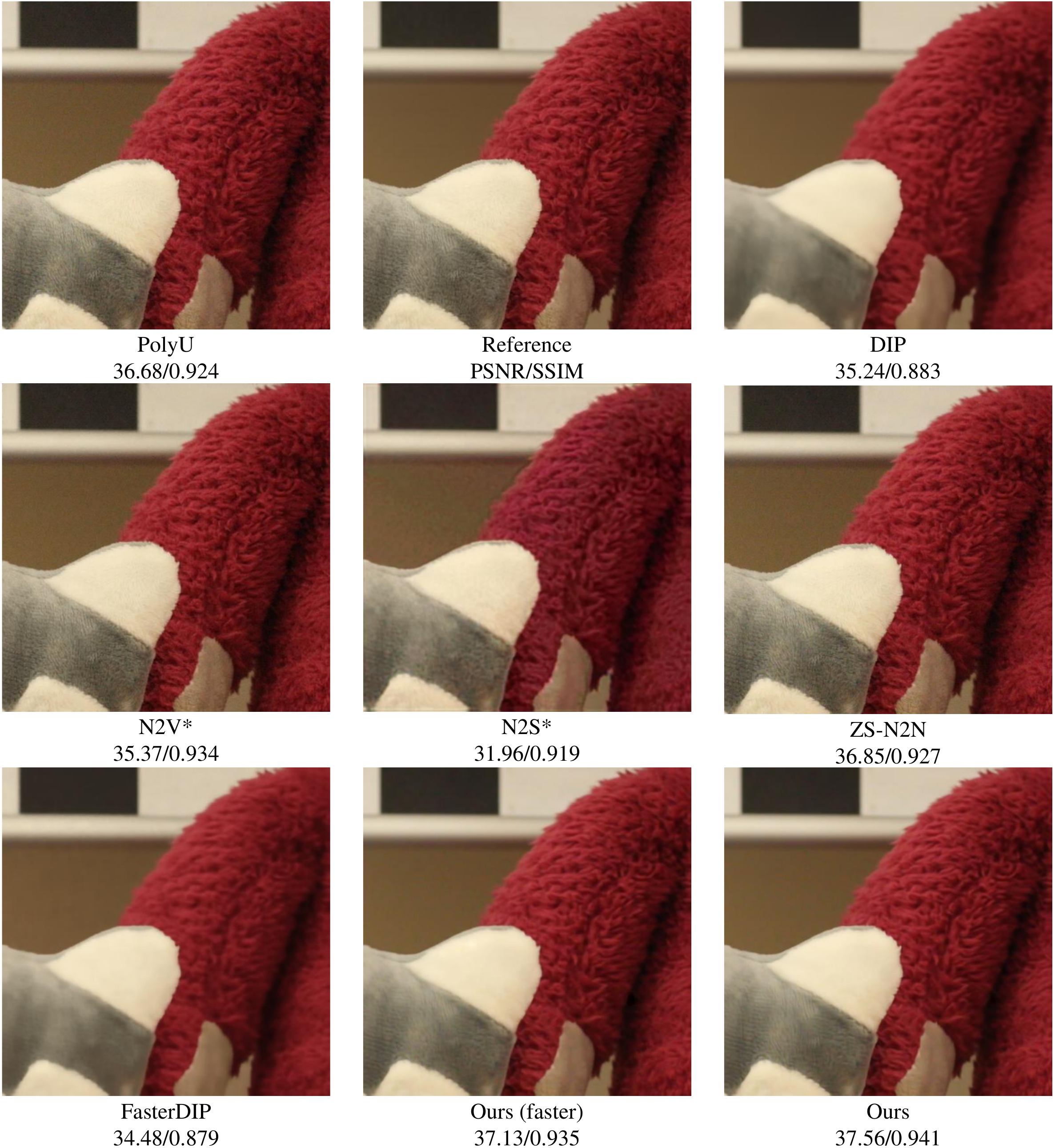}
\end{center}
\caption{Qualitative comparison of realnoise on PolyU~\cite{xu2018polyu}. Noisy patch is from Canon5D2\_5\_200\_3200\_toy\_1.
}
\label{fig_general_polyu2}
\end{figure*}

\begin{figure*}
\begin{center}
\includegraphics[width=1\textwidth]{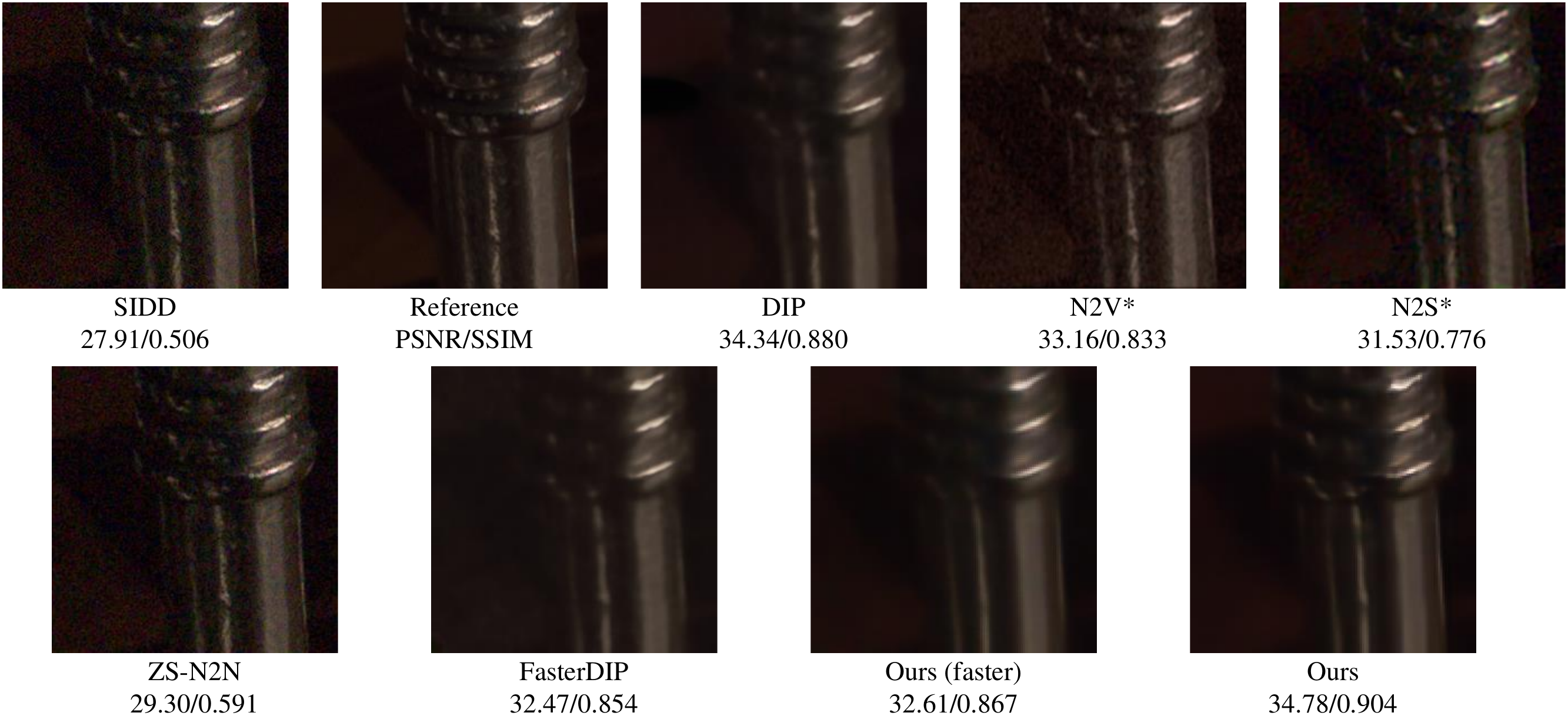}
\end{center}
\caption{Qualitative comparison of realnoise on SIDD~\cite{abdelhamed2018sidd}. Noisy patch is from SIDDval\_20\_8.
}
\label{fig_general_sidd1}
\end{figure*}

\begin{figure*}
\begin{center}
\includegraphics[width=1\textwidth]{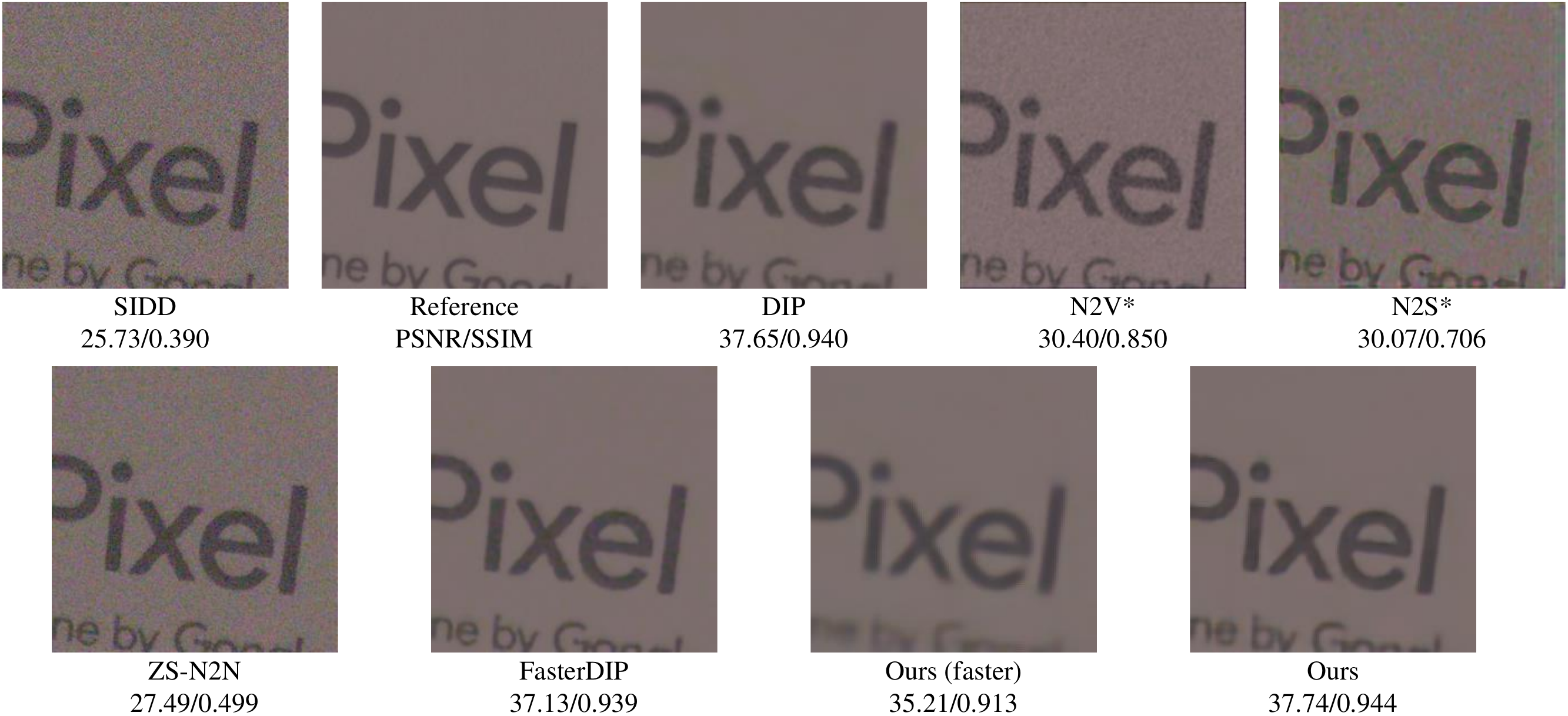}
\end{center}
\caption{Qualitative comparison of realnoise on SIDD~\cite{abdelhamed2018sidd}. Noisy patch is from SIDDval\_13\_19.
}
\label{fig_general_sidd2}
\end{figure*}

The following figures show the denoising comparison on
both synthetic noise removal (see Fig.~\ref{fig_gauss_10} - Fig.~\ref{fig_general_poiss}) and denoising real noise data (Fig.~\ref{fig_general_polyu1} - Fig.~\ref{fig_general_sidd2}).

\end{document}